\documentclass[letterpaper]{article} 
\usepackage{aaai24}   
\usepackage{times}  
\usepackage{helvet}  
\usepackage{courier}  
\usepackage[hyphens]{url}  
\usepackage{graphicx} 
\usepackage{natbib}  
\usepackage{caption} 
\usepackage{algorithm}
\usepackage{algorithmic}

\usepackage{epsfig}
\usepackage{amsmath}
\usepackage{amssymb}
\usepackage{booktabs}
\usepackage[table]{xcolor}
\usepackage[super]{nth}
\usepackage{nth}
\usepackage{multirow}
\usepackage{arydshln}
\usepackage{enumitem}
\usepackage{cuted}
\usepackage{tabularx} 
\usepackage{tabulary}
\newcolumntype{K}[1]{>{\centering\arraybackslash}p{#1}}
\newcolumntype{C}{>{\centering\arraybackslash}X} 
\usepackage{times}
\usepackage{rotating}
\usepackage{xfrac}   
\usepackage{bookmark}

\DeclareMathOperator*{\argmax}{arg\,max}

\DeclareMathOperator*{\range}{range} 
\DeclareMathOperator*{\Lap}{Lap} 
\DeclareMathOperator*{\MLP}{MLP} 
\DeclareMathOperator*{\VED}{VED} 
\DeclareMathOperator*{\concat}{%
    \mathchoice%
        {\Big\Vert}%
        {\big\Vert}%
        {\Vert}%
        {\Vert}%
} 

\usepackage{newfloat}
\usepackage{listings}
\DeclareCaptionStyle{ruled}{labelfont=normalfont,labelsep=colon,strut=off} 
\floatstyle{ruled}
\newfloat{listing}{tb}{lst}{}
\floatname{listing}{Listing}
\title{Disguise without Disruption: Utility-Preserving Face De-Identification}
\author{
    Zikui Cai\textsuperscript{\rm 1 \thanks{This work was primarily carried out during the internship of Zikui Cai at United Imaging Intelligence, Burlington, MA 01803.}}, 
Zhongpai Gao\textsuperscript{\rm 2}, 
Benjamin Planche\textsuperscript{\rm 2}, 
Meng Zheng\textsuperscript{\rm 3}, \\
Terrence Chen\textsuperscript{\rm 2}, 
M. Salman Asif\textsuperscript{\rm 1}, 
Ziyan Wu\textsuperscript{\rm 2}
}
\affiliations{
    \textsuperscript{\rm 1}University of California, Riverside, CA \qquad 
    \texttt{\{zcai032, sasif\}@ucr.edu} \\
    \textsuperscript{\rm 2}United Imaging Intelligence, Burlington, MA \qquad \texttt{\{first.last\}@uii-ai.com}\\
    \textsuperscript{\rm 3}Rensselaer Polytechnic Institute, Troy, NY \qquad \texttt{zhengm5@rpi.edu}
    
}

\usepackage{xspace}
\makeatletter
\DeclareRobustCommand\onedot{\futurelet\@let@token\@onedot}
\def\@onedot{\ifx\@let@token.\else.\null\fi\xspace}
\def\eg{\emph{e.g}\onedot} \def\Eg{\emph{E.g}\onedot}
\def\ie{\emph{i.e}\onedot} 
\def\cf{\emph{c.f}\onedot} 
\def\etc{\emph{etc}\onedot} 
\def\wrt{w.r.t\onedot}

\begin{document}

\maketitle

\begin{abstract}
With the rise of cameras and smart sensors, humanity generates an exponential amount of data. This valuable information, including underrepresented cases like AI in medical settings, can fuel new deep-learning tools. However, data scientists must prioritize ensuring privacy for individuals in these untapped datasets, especially for images or videos with faces, which are prime targets for identification methods. Proposed solutions to de-identify such images often compromise non-identifying facial attributes relevant to downstream tasks.
In this paper, we introduce \textit{Disguise}, a novel algorithm that seamlessly de-identifies facial images while ensuring the usability of the modified data. Unlike previous approaches, our solution is firmly grounded in the domains of differential privacy and ensemble-learning research. Our method involves extracting and substituting depicted identities with synthetic ones, generated using variational mechanisms to maximize obfuscation and non-invertibility. Additionally, we leverage supervision from a mixture-of-experts to disentangle and preserve other utility attributes. We extensively evaluate our method using multiple datasets, demonstrating a higher de-identification rate and superior consistency compared to prior approaches in various downstream tasks.
\end{abstract}


\section{Introduction}
\label{sec:intro}


\begin{figure}[t]
    \centering
    \includegraphics[width=0.45\textwidth]{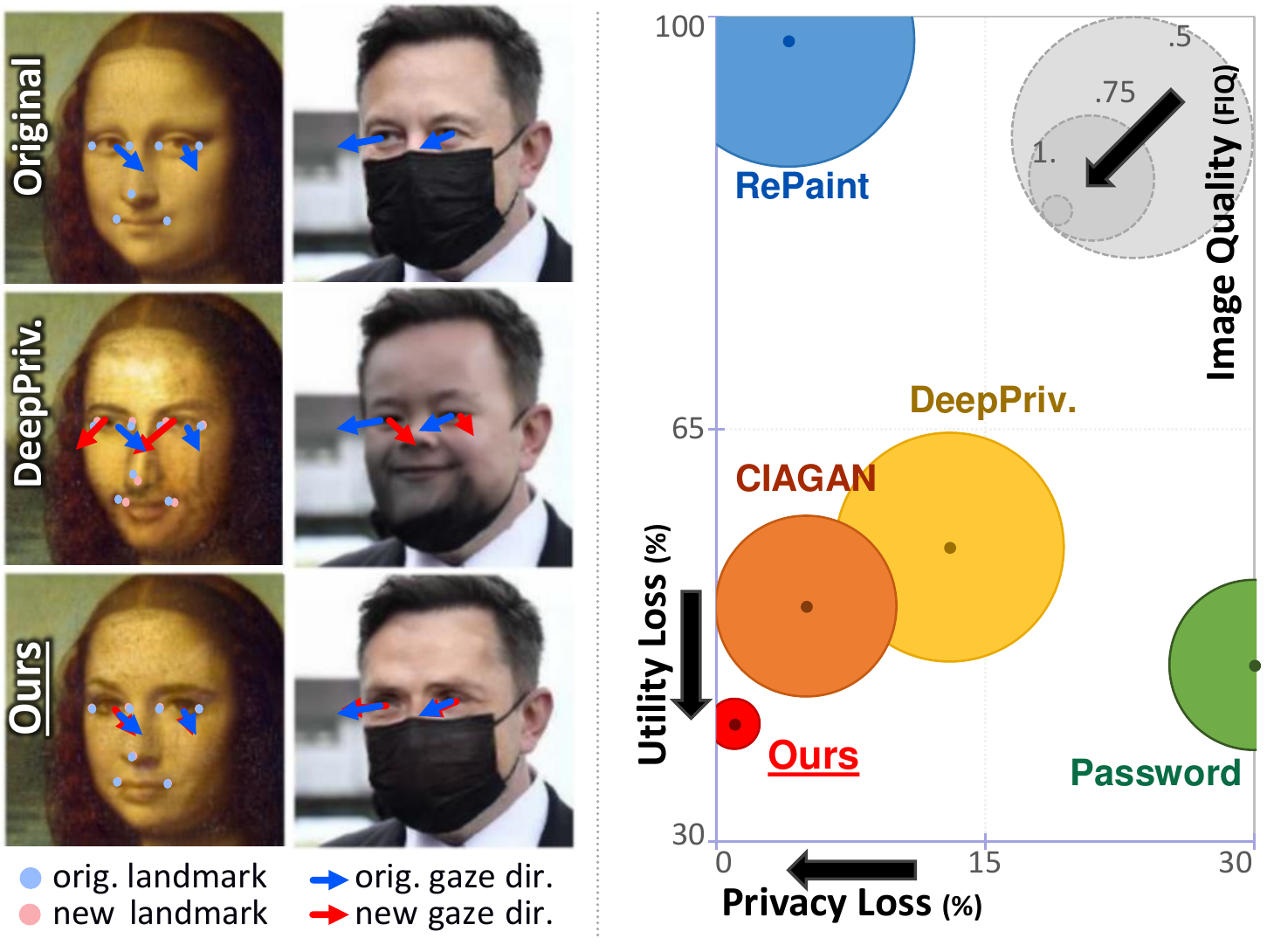}
    \caption{\textit{Disguise} anonymizes face images while preserving their utility (i.e., attributes relevant to downstream tasks). 
    For instance, facial landmarks and gaze direction are better preserved compared to existing methods, as shown in the figure that the red dots for landmarks and red arrows for gazes in the new images are more aligned with the blue ones in the original images. We outperform prior art by a large margin along various axes, including privacy, utility, and image quality. For image quality, small radius indicates higher FIQ \cite{terhorst2020ser} score and better image quality.}
    \label{fig:teaser}\vspace{-10pt}
\end{figure}

Global privacy laws safeguard personal data, including regulations like GDPR \cite{voigt2017eu} in Europe, HIPPA \cite{HIPAA} and CCPA \cite{CCPA} in the US, and PIPL \cite{PIPL} in China. Particularly stringent for medical information and data from medical settings, these rules tightly control storage and distribution of patient images to ensure confidentiality. Yet, this data holds valuable potential, such as automating medical procedures and new AI-driven diagnoses. To tap into these datasets, scientists explore techniques for using sensitive images without compromising identity. Most methods focus on face obfuscation  \cite{newton2005preserving}, blurring \cite{frome2009large}, pixelation \cite{zhou2020personal}, warping \cite{korshunov2013using}), affecting image saliency. Face-swapping  \cite{hukkelaas2019deepprivacy,maximov2020ciagan,chen2020simswap,gu2020password,cao2021personalized,proencca2021uu,agarwal2021privacy} is emerging as a promising solution.

Popularized through the notion of \textit{deepfakes} \cite{westerlund2019emergence}, these deep-learning models are trained to replace any face in an image or video by another one (user-provided or AI-generated), while trying to preserve the overall saliency or specific facial attributes, such as perceived gender, expression, or hair color. While recent solutions can generate convincing results, they are not suitable for the targeted use cases as they lack formal \textit{privacy} and \textit{utility} guarantees for the resulting images. Face-swapping methods evade the confidentiality of the ID provider since the swapped face leaks the source ID. In addition, they lack proper mechanisms to maximize de-identification and minimize identity leakage of the target ID. 
Furthermore, they do not emphasize on maintaining the \textit{utility} of resulting images, \ie, they do not guarantee that the altered images can have the same function as the original ones for various downstream tasks. For example, a dataset would become \textit{useless} for analysis if relevant non-biometric features are corrupted (\eg, facial expressions have changed for sentiment analysis tasks) or for training recognition models if the altered images no longer match their annotations (\eg, facial landmarks, gaze directions, head-pose orientations, \etc). 

In this work, we aim to address the challenge of anonymizing images of individuals while ensuring privacy and maintaining high data utility. 
To this end, we propose \textit{Disguise} (Deep Identity Swapper Guaranteeing Utility with Implicit Supervision from Experts), a de-identification method built upon face-swapping technology that offers formal guarantees regarding identity obfuscation and utility retention. Our main contributions are as follows:

\begin{itemize}
    \item We propose a simple yet effective framework for face de-identification which generates natural faces with distinct identities from the original ones, while maintaining non-biometric attributes unchanged.
    unchanged.
    \item Unlike existing methods that pre-discard original face IDs, we condition the synthetic faces on the original ID vectors and maximize the distance to the original identities while ensuring differential privacy \cite{dwork2014algorithmic,abadi2016deep}, with randomization to prevent re-identification.
    \item We demonstrate superior results than state-of-the-art methods through extensive evaluation regarding the de-identification rate, utility preservation, and image quality of the resulting data over a large number of metrics.
\end{itemize}

\section{Related work}

\noindent\textbf{Face Swapping.} The topic of face swapping has received significant attention in research and is highly relevant, as evidenced by the large body of works dedicated to it  \cite{nirkin2019fsgan,li2019faceshifter,perov2020deepfacelab,chen2020simswap,zhu2021one,xu2022high}. 
However, it presents inherent and important differences compared to face anonymization/de-identification. 
Face swapping aims to change the original identity to a specified target identity, whereas face anonymization shall not rely on actual identities, as it would otherwise compromise both target and source individuals. 
Moreover, the two domains consider different performance indicators and evaluation metrics. Anonymization aims at providing privacy-preserving guarantees, including face anonymization rate and non-re-identifiability \cite{gross2005integrating,liu2021dp,croft2021obfuscation,tolle2022content}, which implies additional mechanisms compared to the face-swapping methods that prioritize preserving facial attributes while reckoning the visual quality of the injected identity \cite{nirkin2019fsgan,xu2022high}.

\noindent\textbf{Face Anonymization.} Although traditional methods such as blacking out, pixelation, and Gaussian blur \cite{boyle2000effects,gross2006model,gross2009face,neustaedter2006blur,newton2005preserving} are effective in removing privacy-sensitive information, they drastically alter the original data distribution, resulting in a significant loss in \textit{utility}. In other words, these methods generate anonymized images that are not suitable for downstream tasks such as gaze estimation \cite{kellnhofer2019gaze360,Zhang2020ETHXGaze}, head-pose prediction \cite{zhou2020whenet,hempel20226d}, facial-landmarks regression \cite{deng2020retinaface,dlib09}, and expression estimation \cite{wen2021distract,savchenko2022video} due to the lack of necessary visual information.

A significant amount of research on face anonymization approaches the problem as an image inpainting task, where the face region is first erased and then replaced with another. 
Early methods \cite{gross2005integrating,padilla2015visual} use a database of real faces to aggregate the new identity, while more recent methods \cite{hukkelaas2019deepprivacy,maximov2020ciagan,liu2021dp} use generative models to synthesize fake identities based on the learned distribution. DeepPrivacy \cite{hukkelaas2019deepprivacy} is one of the pioneering works in this field, which reconstructs the missing face by taking the masked face and facial landmarks as inputs. However, the reconstructed face distribution suffers from bias as it is solely conditioned on its training data, leading to a tendency to generate smiling, young-looking faces. 
CIAGAN \cite{maximov2020ciagan} is another work that uses facial masks and landmarks to generate new faces. 
However, it tends to generate faces with duplicated identities due to the length limitation of the one-hot vector.
RePaint \cite{lugmayr2022repaint}, a recent method based on diffusion models, generates photo-realistic faces with large facial variances, but it fails to maintain the utility of the faces and is sensitive to input distributions.

\begin{figure*}[tb!]
    \centering
    \includegraphics[width=0.9\textwidth]{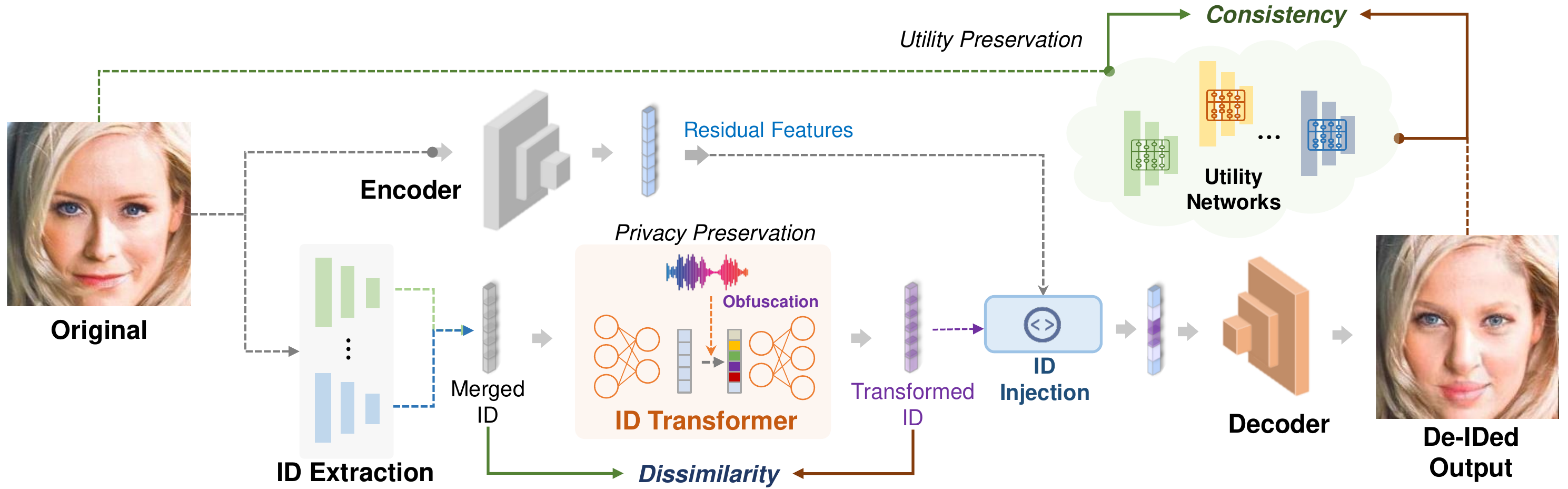}
    \caption{Illustration of the training process for the proposed \textit{Disguise} framework. More discussions in Methodology Section.}
    \label{fig:architecure}
\vspace{-10pt}
\end{figure*}

Some methods \cite{gu2020password,cao2021personalized,proencca2021uu} have focused on making the anonymization process reversible, such as \textit{Password}~\cite{gu2020password} and \textit{RiDDLE}~\cite{li2023riddle}, which generate anonymized faces conditioned on a password that can be used to de-anonymize them. While such a feature can be desirable in some scenarios, it violates privacy regulations like GDPR~\cite{voigt2017eu} that protects \textit{pseudonymous} data (data that has been de-identified from the data's subject but can be re-identified as needed by the use of additional information). In this work, we propose to anonymize faces in an irreversible manner.
Other solutions \cite{li2019anonymousnet,chen2021perceptual,li2021differentially,liu2021dp} incorporate notions of differential privacy \cite{duchi2013local,dwork2014algorithmic,abadi2016deep} by adding adequately-calibrated random noise either at training or inference time, ensuring privacy levels linked to their parameter $\epsilon$. Or directly optimize in the latent space of StyleGAN \cite{barattin2023attribute}. However, they often neglect utility preservation (\eg, they edit image background and utility attributes) and require complex post-processing, making them not readily applicable to anonymization tasks.




\section{Methodology}
\label{sec:method}

In this section, we formalize our objectives, theoretically ground our work, and finally describe our proposed solution.


\subsection{Problem Formulation}
\label{sec:formalism}

\noindent\textbf{Privacy Utility Dual Optimization.}
Let $\mathcal{X} \subset \mathbb{R}^{3\times H \times W}$ be the image space
, with $x \in \mathcal{X}$ an image depicting an individual.
Let $\left(\mathcal{Z}, d_\mathcal{Z}\right)$ be a metric space, with $\mathcal{Z} \subset \mathbb{R}^{n_\mathcal{Z}}$ space of identity-distilled facial features 
(\ie, facial features that uniquely identify an individual) and $d_\mathcal{Z}: \mathcal{Z} \times \mathcal{Z} \rightarrow \mathbb{R}$ a distance function attached to space $\mathcal{Z}$.
Let $\left(\mathcal{Y}, d_\mathcal{Y}\right)$ be another metric space, with $\mathcal{Y} \subset \mathbb{R}^{n_\mathcal{Y}}$ containing utility-distilled facial features 
(\ie, features that are useful to downstream tasks) and $d_\mathcal{Y}: \mathcal{Y} \times \mathcal{Y} \rightarrow \mathbb{R}$ a distance function relating to $\mathcal{Y}$.
We note $f_\mathcal{Z}: \mathcal{X} \rightarrow \mathcal{Z}$ and $f_\mathcal{Y}: \mathcal{X} \rightarrow \mathcal{Y}$ the objective labeling functions respective to each domain. 


We define a conditional generative function $G: \mathcal{X} \rightarrow \mathcal{X}$ parameterized by $\theta$, that takes $x \in \mathcal{X}$ as input and returns an edited version $G(x) = \widetilde{x}$. Our goal is to learn a $G$ such that \textit{utility} is maximized (\ie, $f_\mathcal{Y}(x) = f_\mathcal{Y}(\widetilde{x})$) and \textit{privacy} is maximized (\ie, $f_\mathcal{Z}(x)$ is distant from $ f_\mathcal{Z}(\widetilde{x})$). In other terms, the output of $G$ should contain the same utility attributes as the input and contain identity attributes different from the input beyond recognition.
Formally, we want $G$ to achieve Pareto optimality \cite{sener2018multi,momma2022multi} \wrt the aforementioned multiple objectives (\ie, identity obfuscation and utility preservation), accounting for their possible competition (depending on downstream tasks, utility and identity attributes may overlap), thus minimizing the following objective:
\vspace{-3pt}
\begin{equation}
\begin{aligned}
    \min_{\theta} 
        \Bigl(
            -&\mathbb{E}_{x \in \mathcal{X}}\left[
                d_\mathcal{Z}\bigl(f_\mathcal{Z}\left(x\right), f_\mathcal{Z} \circ G_\theta\left(x\right)\bigr)
            \right], \\ 
            &\mathbb{E}_{x \in \mathcal{X}}\left[
                d_\mathcal{Y}\bigl(f_\mathcal{Y}\left(x\right), f_\mathcal{Y} \circ G_\theta\left(x\right)\bigr)
            \right]
        \Bigr)^\intercal
\end{aligned}
\end{equation}
Before tackling the challenges of multi-objective optimization that such a task brings, one has to consider how to model the unknown objective distance and labeling functions $d_\mathcal{Z}, f_\mathcal{Z}$ and $d_\mathcal{Y}, f_\mathcal{Y}$ for the identity and utility space respectively.
We argue that identity and utility are conceptually subjective, \ie, different authoritative entities have different definitions and target features assigned to each concept. 
\eg, given a picture of a person, each human or algorithmic agent will rely on different features (facial landmarks, eye color, \etc) and their own subjective judgment to certify the person's identity, as there is no absolute objective function to perform the ill-posed mapping of a facial picture to an identity. Similarly, the concept of \textit{utility} is conditioned by a set of target tasks or the agents in charge of said tasks. 
\eg, an image with the person's face completely blurred could still be \textit{used} by a person-detection algorithm, but would be \textit{useless} for facial landmark regression.

Therefore, we propose to rely on predefined agents (\textit{experts}) to provide the identity and utility definitions to guide the optimization of our model \cite{gross2005integrating}.
We thus consider some parameterized models $h_\mathcal{Z}$ and  $h_\mathcal{Y}$ pre-optimized to approximate their respective objective labeling functions $f_\mathcal{Z}$ and $f_\mathcal{Y}$. Note that we make no assumption on the architecture or training of each of these models (we demonstrate with various state-of-the-art identity extraction and recognition models). 
Without loss of generality and to account for individual bias, we define $H_\mathcal{Z} = \left\{h_\mathcal{Z}^i\right\}_{i=1}^{k_\mathcal{Z}}$ and $H_\mathcal{Y} = \left\{h_\mathcal{Y}^i\right\}_{i=1}^{k_\mathcal{Y}}$ as sets of $k_{{\mathcal{Z}}}$ and $k_{{\mathcal{Y}}}$ unique models which differ in terms of architecture and/or training regime, \cf mixture-of-experts theory \cite{miller1996mixture,masoudnia2014mixture,dai2021generalizable}.
We demonstrate in this paper how these identification/utilization experts can be leveraged in an adversarial/collaborative framework to train $g$ towards a satisfying optimum.

\vspace{.5em}
\noindent\textbf{Identity Obfuscation Guarantees.}
To provide formal de-identification guarantees, we ground our work in the extensive theory on $\epsilon$-differential privacy ($\epsilon$-DP) and $\epsilon$-local-differential privacy ($\epsilon$-LDP, relevant when obfuscation should be performed without global knowledge) applied to identity-swapping functions \cite{duchi2013local,dwork2014algorithmic,abadi2016deep,yu2020gan,liu2021dp,croft2021obfuscation,tolle2022content,qiu2022novel}. 
Let $\psi: \mathcal{Z} \rightarrow \mathcal{Z}$ be a function that performs ID obfuscation, 
\ie, taking an identity vector $z$ and returning a new one $\widetilde{z}$ that maximizes $d_\mathcal{Z}(z, \widetilde{z})$.
We consider that an approximate but randomized function $\psi^\epsilon: \mathcal{Z} \rightarrow \mathcal{Z}$ satisfies 
$\epsilon$-LDP
if, for any two adjacent 
inputs $z$, $z' \in \mathcal{Z}$
and for any subset of outputs $Z_s \subseteq \range(\psi^\epsilon)$, it holds that $\mathrm{P}\left(\psi^\epsilon(z)\right) \leq e^\epsilon \mathrm{P}\left(\psi^\epsilon(z')\right)$.
Given $\Delta\psi = \sup_{z, z' \in \mathcal{Z}}\left\|\psi(z) - \psi(z')\right\|_1$ the sensitivity of $\psi$, Laplace noise is commonly leveraged to define an $\epsilon$-DP version of the function: $\psi^\epsilon(z) \triangleq \psi(z) + \left(\Lap\left({\sfrac{\Delta\psi}{\epsilon}}\right)\right)^{n_\mathcal{Z}}$ \cite{duchi2013local,dwork2014algorithmic,abadi2016deep,liu2021dp}.
We demonstrate that to ensure $\epsilon$-LDP, the $d_\mathcal{Z}$-maximization property of the identity-obfuscation function has to be relaxed. The manifold of identity vectors generated by an identification function $h_{\mathcal{Z}}$ is bounded by the range of said function. In such a space and for any Euclidean distance $d_\mathcal{Z}$, a non-relaxed version of $\psi$ would be the bijective (and thus non-private) function $\psi_{\text{opp}}$ mapping an ID vector to its opposite. No other function (\eg, $\psi^\epsilon$) could ensure $d_\mathcal{Z}$-maximization.
Therefore, in this work, we consider the inherent trade-off between maximizing swapping-based identity obfuscation and ensuring differential privacy, and we propose a variety of solutions $\psi^\epsilon$ tailored to different needs (as illustrated in Figure~\ref{fig:transform}, and more details in Proposed Solution Section).

\noindent\textbf{Non Re-identifiability.}
Another important aspect to consider in privacy-preserving applications is \textit{non-invertibility}. If the de-identified data can be re-identified with additional information, then the operation is not truly anonymization but \textit{pseudonymization}. For example, with the correct password for Password \cite{gu2020password} and RiDDLE~\cite{li2023riddle}, or using the opposite ID for $\psi_{\text{opp}}$, the original ID is compromized.
We empirically demonstrate that the proposed obfuscation solutions achieve varying degrees of robustness to such re-identification efforts. 

In the remaining of the section, we explain how we define and train $g$ to ensure privacy-preserving non-invertible identity swapping in images and utility preservation.

\subsection{Proposed Solution}\label{sec:solution}

The proposed architecture can be defined as the composition of a face-swapping model $g: \mathcal{X} \times \mathcal{Z} \rightarrow \mathcal{X}$, an identity extractor $h_{\mathcal{Z}}: \mathcal{X} \rightarrow \mathcal{Z}$, and an identity obfuscation function $\psi^\epsilon: \mathcal{Z} \rightarrow \mathcal{Z}$, such that $G(x) = g\left(x, \psi^\epsilon \circ h_{\mathcal{Z}}(x)\right)$. 
Given a facial image $x$, $h_{\mathcal{Z}}$ extracts the vector $z$ encoding the identity of the depicted person. This vector $z$ is passed to the privacy-enabling function $\psi^\epsilon$, which returns a synthetic identity $\widetilde{z}$ that maximizes obfuscation. 
Finally, the face-swapper model $g$ edits the original image $x$ to inject the fake identity $\widetilde{z}$, resulting in an image $\widetilde{x}$ where the original visual identifying attributes are replaced by those encoded in $\widetilde{z}$. 
Additionally, during its training, $g$ relies on the feedback of tasks-specific models $h_{\mathcal{Y}}^i: \mathcal{X} \rightarrow \mathcal{Y}$ to ensure that the utility of $\widetilde{x}$ is maintained compared to $x$.
We expand on each block in the following paragraphs.

\begin{figure}[tb!]
    \centering
    \includegraphics[width=0.47\textwidth]{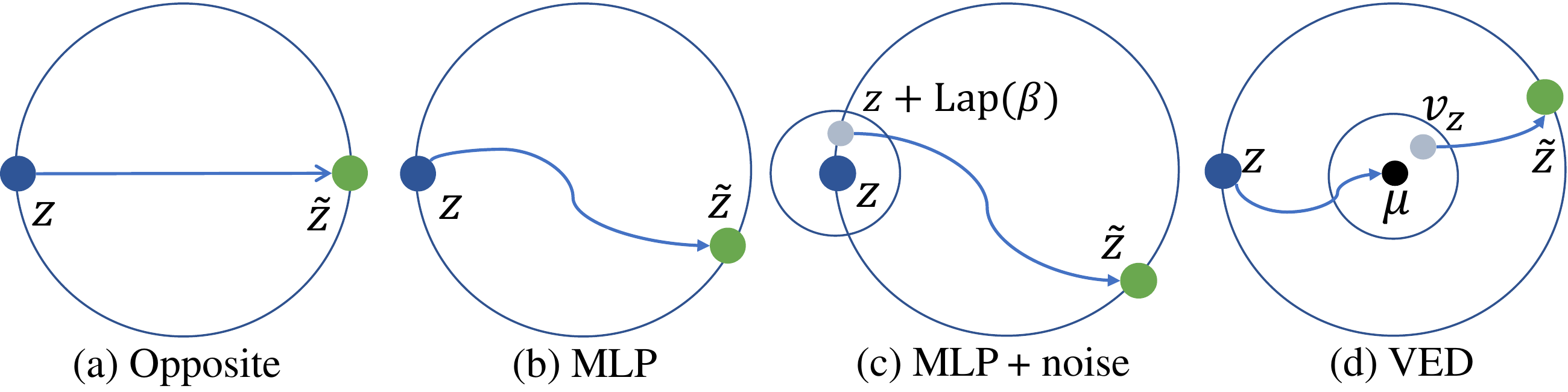}
\caption{Identity transformation. The identity vector is normalized to the surface of a unit n-sphere.}
\label{fig:transform}\vspace{-12pt}
\end{figure}


\vspace{.2em}
\noindent\textbf{
Identity Extraction.} 
As mentioned in Section~\ref{sec:formalism}, we propose to extract the identity information from facial images via model ensembling \cite{miller1996mixture,masoudnia2014mixture,dai2021generalizable}, to ensure generalizability 
as well as to limit the impact of models' bias (as we assume no control over the architecture or training regimen of selected identity-expert models).
Therefore, given a set $H_\mathcal{Z} = \left\{h_\mathcal{Z}^i\right\}_{i=1}^{k_\mathcal{Z}}$ of ID extractors, we define $h_{\mathcal{Z}}$ as the ensemble method $h_{\mathcal{Z}}(x) = \MLP_{\theta_z}\big(\concat_{i=1}^{k_\mathcal{Z}}h_\mathcal{Z}^i(x)\big)$, \ie, concatenating (symbol $\Vert$) the $k_\mathcal{Z}$ predicted vectors together and merging them into $z \in \mathcal{Z}$ via a multilayer perceptron (MLP) with parameters $\theta_z$ and $\tanh$ as final activation. 


\vspace{.2em}
\noindent\textbf{
Identity Transformation.}
\label{sec:id_trans}
A variety of techniques can be considered to perform the identity transformation $\psi$, as shown in Figure~\ref{fig:transform}. 
If we were to maximize the distance between the original and obfuscating IDs, the optimal function would be $\psi_{\text{opp}}(z) = -z$ since $\argmax_{z'}d_\mathcal{Z}(z, z') = -z$  in our normalized Euclidean identity space. 
However, such a function is reversible, making it easy to re-identify the original individual by taking the opposite of the pseudonymized ID.
A more secure solution would be a parametric function, \eg, $\psi_{\text{mlp}}(z) = \MLP_{\theta_{\psi}}(z)$, trained to optimally fool $h_{\mathcal{Z}}$.  
As a non-explicit function, $\psi_{\text{mlp}}$ is more challenging to invert, though not impossible with the access to the model or its parameters $\theta_{\psi}$ (\cf gradient-based attacks \cite{fredrikson2015model,wang2015regression}). 
To increase robustness and ensure $\epsilon$-LDP, we can add dimension-wise noise to the inner operation, \ie, $\psi^\epsilon_{\text{mlp}}(z) = \MLP_{\theta_{\psi}}\left(z + \left(\Lap\left(\beta\right)\right)^{n_\mathcal{Z}}\right)$, with $\beta = {\frac{\Delta\psi_{\text{mlp}}}{\epsilon}}$.
The larger $\beta$ is set (\ie, the smaller $\epsilon$ is), the more noise is applied to the original ID vector before further MLP-based obfuscation. Therefore, larger $\beta$ provides stricter privacy guarantee and robustness but adversarial affects the ability of $\psi^\epsilon_{\text{mlp}}$ to learn how to fool identification experts $H_\mathcal{Z}$.

To better navigate this trade-off and guarantee a more continuous space for the noise application, we leverage the properties inherent to variational autoencoders (VAEs) \cite{kingma2013auto}. We introduce a variational encoder-decoder (VED) to transform the identity vector, \ie,
$\psi^\epsilon_{\text{ved}}(z) = \VED_{\theta_{\psi'}}\left(z\right)$. 
This model's encoder predicts the parameters $(\mu, \sigma)$ of the latent data distribution (assumed to be Gaussian). 
A latent vector $v_z$ is picked as $\mu + \sigma \eta$ with $\eta \sim \left(\mathcal{N}(0, 1)\right)^{n_v}$ (\cf reparameterization \cite{kingma2015variational}) then passed to the decoder. While a VAE decoder would reconstruct the input identity from $v_z$, our VED decoder should generate a new, distant identity.
During inference, we sample $v_z$ as $\mu + \sigma \left(\Lap\left(\alpha\right)\right)^{n_v}$ to meet $\epsilon$-LDP, with ${n_v}$ dimension of latent space and $\alpha = {\frac{\Delta\psi_{\text{ved}}}{\epsilon}}$. To train either of these models, we enforce cosine dissimilarity between the original and generated ID vectors:

\begin{equation}
    \mathcal{L}_{\text{deid}} = 1 + \frac{z \cdot \widetilde{z}}{\|z\|_2\|\widetilde{z}\|_2}.
\end{equation}
For the VED model, we add to this criterion the usual Kullback–Leibler divergence (KLD) loss $\mathcal{L}_{\text{kld}}$ \cite{kingma2015variational,kingma2013auto}.

\vspace{.2em}
\noindent\textbf{
Face Swapping.}
Once the fake identity vector $\widetilde{z}$ is generated, it is passed to the face-swapping model $g$, along with the original image $x$. 
Similar to existing solutions \cite{chen2020simswap,perov2020deepfacelab,liu2021dp}, $g$ is composed of three modules: (1) an image encoder that extracts identity-unrelated features $\nu$; (2) an ID injector that aggregates $\nu$ and $\widetilde{z}$ into a vector encoding the content of the obfuscated image $\widetilde{x}$; (3) a decoder conditioned on this vector that generates $\widetilde{x}$. 
These existing works also share similar losses that we borrow and adapt:

\vspace{-5pt}
\begin{equation}
    \begin{aligned}
        \mathcal{L}_{\text{mix}} &= \|g(x, \widetilde{z}) - g(x, z)\|_1 ; 
        \\
        \mathcal{L}_{\text{gen}} &= \sum_{i=1}^{k_d}\log\left(1 - D_i(x, \widetilde{x})\right) ;
        \\
        \mathcal{L}_{\text{id}} &= \sum_{\hat{z} \in \{z, \widetilde{z}\}}
            \Big(
            1 - \frac{\hat{z} \cdot \hat{z}_h}{\|\hat{z}\|_2\|\hat{z}_h\|_2}
            \Big)
            ;
    \end{aligned}
\end{equation} 
with $\hat{z}_h = h_\mathcal{Z}\left(g(x, \hat{z})\right)$.
Here, $\mathcal{L}_{\text{mix}}$ is a mixing loss to ensure implicit disentanglement of ID features (encoded in $z$ or $\widetilde{z}$) and residual features (\ie, $\nu$).
$\mathcal{L}_{\text{gen}}$ pits the generator against $k_d$ discriminators $D$ to ensure realistic results preserving image saliency, \cf recent GAN solutions \cite{wang2018pix2pixHD,hukkelaas2019deepprivacy,chen2020simswap} (we also use their weak-feature matching loss, further ensuring the high-level semantic alignment between the image pairs).
Finally, $\mathcal{L}_{\text{id}}$ enforces cosine similarity between the injected identity $\hat{z}$ and the one observed by the identification model $h_\mathcal{Z}$ in the resulting image.
Combined together, along with $\mathcal{L}_{\text{deid}}$ and $\mathcal{L}_{\text{kld}}$ (using weighting hyperparameters), these losses form the overall objective for our privacy-enforcing face-swapping solution $G$.


\begin{figure}[t!]
    \centering
    \includegraphics[width=.47\textwidth]{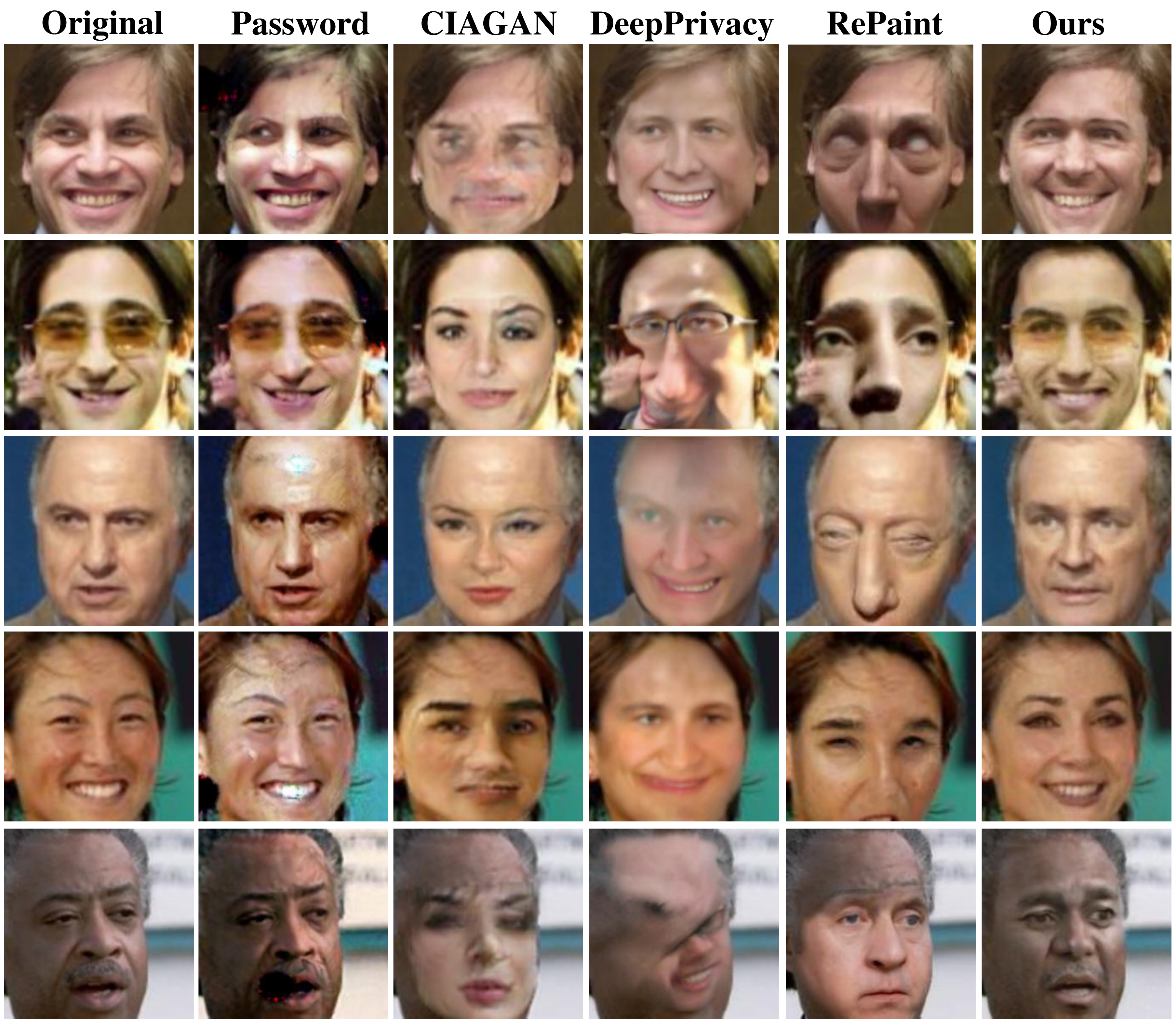}
    \caption{Qualitative results of different methods. Ours preserves utility while anonymizing identities.}
    \label{fig:qualities}\vspace{-10pt}
\end{figure}

\vspace{.5em}
\noindent\textbf{
Utility Preservation.}
Existing face-swapping methods \cite{hukkelaas2019deepprivacy,chen2020simswap,perov2020deepfacelab,liu2021dp} claim that their adversarial and feature-matching losses ensure the preservation of non-identifying content. However, such supervision is too weak to guarantee that the images will maintain their utility \wrt downstream tasks, especially for tasks relying on small attention regions (\eg, gaze estimation). 
We thus complement the aforementioned objective with a criterion that leverages the implicit expertise of tasks-relevant models $H_\mathcal{Y}$:

\begin{equation}\label{eq:uti_loss}
    \mathcal{L}_{\text{uti}} = \sum_{i=1}^{k_\mathcal{Y}} \lambda_{\text{uti},i} \| h_\mathcal{Y}^{i,l}(x) - h_\mathcal{Y}^{i,l}(\widetilde{x}) \|_1,
\end{equation}
with $h_\mathcal{Y}^{i,l}(\boldsymbol{\cdot})$ the features returned by the last differential non-softmax layer $l$ of model $h_\mathcal{Y}^i$, and $\lambda_{\text{uti}} \in \mathbb{R}^{k_\mathcal{Y}}$ hyperparameters weighting the task/expert contributions. 
Hence, $\mathcal{L}_{\text{uti}}$ imposes that altered images contains the same utility attributes as original images, as expected by tasks-relevant models.

Note that the entire solution $G(x) = g\left(x, \psi^\epsilon \circ h_{\mathcal{Z}}(x)\right)$ is end-to-end differentiable, thus single-pass trainable. 
In practice, we leverage its modularity and train each component separately before jointly fine-tuning.
Scalar hyperparameters weigh the contribution of each loss to the total objective (we fix $\{\lambda_{\text{id}}, \lambda_{\text{deid}}, \lambda_{\text{mix}}, \lambda_{\text{uti,eye}}, \lambda_{\text{uti,emo}}, \lambda_{\text{kld}}\} = \{30, 30, 10, 2, 2, 0.2\}$).
\section{Experiments}\label{sec:exp}



\begin{figure*}[t]
    \centering
    \includegraphics[width=1\linewidth]{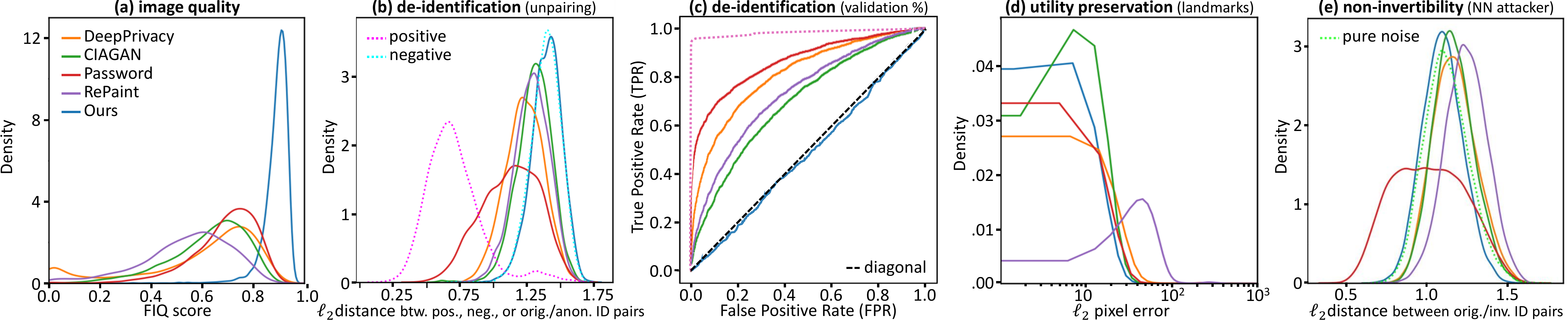}
    \vspace{-12pt}
    \caption{\textit{Disguise} outperforms existing methods in various aspects, including image quality, de-id rate, and utility. For non-invertibility, our solution is close to other methods that completely erase the original IDs (\ie, recovering pure Gaussian noise).}
    \label{fig:hist-4dist}
    \vspace{-10pt}
\end{figure*}

We now describe our experimental setup and compare with other methods in terms of privacy robustness and data usability. More details in supplementary material.

\subsection{Experimental Protocol}\label{sec:protocol}




\noindent\textbf{Datasets.} We use multiple datasets for taining and evaluation. We train our models on VGGFace2 dataset \cite{Cao18}, which totals 3.31 million images with 9,131 identities. We use multiple datasets for evaluation, including 
LFW \cite{huang2008labeled} (13,233 face images and 5,749 identities) for utility and de-identification performance, CelebA-HQ \cite{karras2017progressive} (30,000 face images) for utility evaluation, and WFLW \cite{wayne2018lab} (10,000 face images) for the training usability \wrt the  downstream task of landmark detection.


\noindent\textbf{Identity and Utility Models.} To demonstrate the genericity of our method, we consider a variety of pretrained face-identification networksand of utility networks over different recognition tasks.
As identity experts, we use ArcFace \cite{deng2019arcface}, AdaFace \cite{kim2022adaface}, FaceNet \cite{schroff2015facenet}, and SphereFace \cite{Liu_2017_CVPR}. 
Either ArcFace ($h_\mathcal{Z}^\text{arc}$), AdaFace ($h_\mathcal{Z}^\text{ada}$), or both ($h_\mathcal{Z}^\text{mix}$) are used to guide $g$ during training (\cf Equation \ref{eq:uti_loss}); FaceNet and SphereFace are used only for evaluation. 
For the downstream tasks, we use ETH-XGaze \cite{Zhang2020ETHXGaze} (noted $h_\mathcal{Y}^\text{eye}$) for gaze estimation, DAN \cite{wen2021distract} ($h_\mathcal{Y}^\text{emo}$) for facial expression recognition, or both ($h_\mathcal{Y}^\text{mix}$) to provide utility feedback during training.
During evaluation, we use L2CS-Net \cite{abdelrahman2022l2cs} for gaze estimation, DeepFace (DF) \cite{serengil2021lightface} for emotion recognition, and
RetinaFace \cite{deng2020retinaface} 
and Dlib \cite{dlib09} for landmark detection. 

\noindent\textbf{
Metrics.} We employ the commonly-used validation rate and verification accuracy as metrics for evaluating privacy preservability \cite{schroff2015facenet,Liu_2017_CVPR,deng2019arcface,kim2022adaface}. The validation rate is defined as the true positive rate (TPR) at certain false positive rate (FPR), \eg, TPR @ FPR=1e-3. 
Verification accuracy is the percentage of image pairs correctly classified as the same/different person using the best $\ell_2$ distance threshold. The verification accuracy of random guessing is thus 50\%, which is what anonymization aims at. 
To measure utility preservation, we use $\ell_2$ pixel distance and normalized mean error (NME) for facial landmark detection, mean absolute error (MAE) for gaze estimation, and accuracy for emotion recognition. For image quality, we use SER-FIQ \cite{terhorst2020ser}.

\noindent\textbf{Comparison.} We consider various de-identification methods, including DeepPrivacy \cite{hukkelaas2019deepprivacy}, DeepPrivacy2 \cite{hukkelaas2023deepprivacy2}, CIAGAN \cite{maximov2020ciagan}, Password \cite{gu2020password}, and RePaint \cite{lugmayr2022repaint}.
For readibility of the tables, we denote different versions as ``Ours ($a$, $b$, $c$)" where $a$ fixes the identity model(s) $h_\mathcal{Z}^a$ used, $b$ the transformation function $\psi_b^\epsilon$, and $c$ the utility model(s) $h_\mathcal{Y}^c$.
For simplicity, we use ``Ours ($\text{arc}, \text{ved}, \text{eye}$)" as our default method unless otherwise mentioned. We demonstrate the impact of different transformation models and identity/utility experts in ablation studies.



\subsection{Privacy: Obfuscation Evaluation}
\label{sec:exp-privacy}

\noindent\textbf{De-identification performance.}
As shown in Table \ref{tab:compare-deid}, we achieve near perfect de-id rate, \ie, with a validation rate close to $0$ and verification accuracy close to $50\%$, outperforming other methods by a significant margin, and is even more secure than randomly picking replacement images from the dataset. Figure \ref{fig:hist-4dist}(b) presents the $\ell_2$ distance histogram for original positive pairs, original negative pairs, and original-anonymized positive pairs on LFW \cite{huang2008labeled}, and Figure \ref{fig:hist-4dist}(c) shows the ROC curves of validation rate. We observe that \textit{Disguise} creates image pairs that are close to the negative distribution, hence perfect obfuscation. We also achieve the highest facial image quality, see Figures \ref{fig:teaser} and \ref{fig:qualities} for visual reference. Among other comparing methods, it is worth noticing that Password fails to de-identify images, hence the highest validation rate. CIAGAN and RePaint are better than Password in de-identification, however they suffer from low facial image quality due to high artifacts and distortions.

\begin{table}[t]
\centering
\small
\caption{Identification / validation rate and image quality evaluation over edited LFW data.}
\label{tab:compare-deid}
\vspace{-6pt}

\resizebox{\linewidth}{!}{
\begin{tabular}{l@{\hspace{0.2em}}l@{\hspace{0.15em}}l@{\hspace{0.15em}}r|cccc|c}
\toprule
\multicolumn{4}{c|}{\multirow{2}{*}{Methods}} & \multicolumn{4}{c|}{TPR (\%) @ FPR=$10^{-3}$ / Accuracy (\%) $\downarrow$} & FIQ $\uparrow$ \\
\multicolumn{4}{c|}{} & FaceNet & SphereFace & AdaFace & Average & SER \\ \midrule
\multicolumn{4}{l|}{Original} & 93.8 / 97.1 & 87.9 / 96.2 & 95.4 / 97.7 & 92.4 / 97.0 & 0.77 \\
\multicolumn{4}{l|}{DeepPrivacy} & 7.3 / 73.8 & 2.9 / 70.9 & 4.6 / 68.6 & 4.9 / 71.1 & 0.67 \\
\multicolumn{4}{l|}{DeepPrivacy2} & 1.7 / 62.5 & 1.0 / 61.5 & 2.2 / 62.2 & 1.6 / 62.1 & 0.68 \\
\multicolumn{4}{l|}{CIAGAN} & 1.8 / 64.5 & 1.0 / 59.0 & 5.6 / 71.0 & 2.8 / 64.8 & 0.58 \\
\multicolumn{4}{l|}{Password} & 31.7 / 79.1 & 17.1 / 73.5 & 51.0 / 84.0 & 33.3 / 78.9 & 0.69 \\
\multicolumn{4}{l|}{RePaint} & 2.8 / 67.7 & 1.1 / 63.5 & 3.6 / 68.5 & 2.5 / 66.6 & 0.54 \\ 
\multicolumn{4}{l|}{Ours} 
& 
\textbf{0.03} / \textbf{50.0} & \textbf{0.03} / \textbf{50.0} & \textbf{0.00} / \textbf{50.0} & \textbf{0.02} / \textbf{50.0} & \textbf{0.90} \\
\bottomrule
\end{tabular}
}
\vspace{-10pt}
\end{table}

\noindent\textbf{Original and anonymized ID de-correlation.} 
We consider scenarios where malicious attackers attempt to link anonymized IDs with their original IDs, allowing them to perform inversion inference on the anonymized IDs and recover the original ones. We use encoder-decoder networks to learn the correlation on existing original-anonymized image pairs. Figure \ref{fig:hist-4dist}(e) shows the results of using MLPs to decode obfuscated IDs from CelebA-HQ \cite{karras2017progressive} while trained on LFW \cite{huang2008labeled} using original IDs as supervision. 
While methods like DeepPrivacy, CIAGAN, and RePaint are inherently robust to inversion attacks since the original face region is entirely erased, and their networks are solely tasked with inpainting the blank region, our method still offers de-correlation on par with these methods, suggesting that our method is also resilient to inversion attacks.

\subsection{Utility: Usability Evaluation}
\begin{table*}[t]
\vspace{-12pt}
\centering
\small
\caption{Utility performance comparison of different anonymization methods over diverse downstream tasks on LFW and CelebA-HQ datasets \cite{huang2008labeled,karras2017progressive}.}
\label{tab:compare-utility}
\vspace{-6pt}
\resizebox{\textwidth}{!}{
\begin{tabular}{l@{\hspace{0.4em}}|l@{\hspace{0.2em}}l@{\hspace{0.15em}}l@{\hspace{0.15em}}r|cccc|cccc|cc|cc|cc}
\toprule
\multicolumn{1}{c|}{\multirow{3}[1]{*}{\begin{sideways}Dataset\end{sideways}} } & \multicolumn{4}{c|}{\multirow{3}{*}{Methods}} & \multicolumn{8}{c|}{Facial landmarks  (L2 pixel distance $\downarrow$)} & \multicolumn{4}{c|}{\begin{tabular}[c]{@{}c@{}}Gaze estimation  (MAE° $\downarrow$)\end{tabular}} & \multicolumn{2}{c}{\begin{tabular}[c]{@{}c@{}}Emotion \end{tabular}} \\ \cline{6-19} 
& \multicolumn{4}{c|}{} & \multicolumn{4}{c|}{RetinaFace (5 points)} & \multicolumn{4}{c|}{Dlib (68 points)} & \multicolumn{2}{c|}{L2CS-Net} & \multicolumn{2}{c|}{ETH-XGaze} & \multicolumn{2}{c}{(Accuracy \% $\uparrow$)}\\
& \multicolumn{4}{c|}{}  & All & Eyes & Nose & Mouth & All & Eyes & Nose & Mouth & Pitch & Yaw & Pitch & Yaw & DAN & DF \\ 
\midrule
\multirow{5}[1]{*}{\begin{sideways}\footnotesize LFW \end{sideways}} 
&
\multicolumn{4}{l|}{DeepPrivacy} & 23.9 & 13.1 & 9.9 & 16.5 & 263.0 & 32.7 & 25.1 & 89.0 & 7.7 & 13.6 & 8.0 & 16.3 & 27.1 & 34.3 \\
& \multicolumn{4}{l|}{DeepPrivacy2} & 31.2 & 18.4  & 14.4  & 19.6 & 385.7  & 59.9  & 49.9  & 120.6 & 9.2 & 12.2 & 7.8  & 15.1 & 22.4 & 30.2 \\
& \multicolumn{4}{l|}{CIAGAN} & 14.6 & 9.3 & 5.5 & 9.2 & 348.2 & 59.0 & 31.2 & 97.7 & 8.8 & 14.6 & 7.8 & 16.9 & 32.5 & 36.9 \\
& \multicolumn{4}{l|}{Password} & 17.4 & 10.4 & 7.7 & 11.1 & 204.8 & \textbf{26.5} & \textbf{19.3} & \textbf{55.4} & 10.5 & 24.7 & 7.7 & 11.5 & 45.9 & 43.4 \\
& \multicolumn{4}{l|}{RePaint} & 66.1 & 30.8 & 32.2 & 47.3 & 1103.1 & 133.5 & 152.1 & 432.0 & 11.3 & 18.1 & 9.2 & 18.8 & 17.3 & 19.4 \\
& \multicolumn{4}{l|}{Ours} 
& \textbf{12.9} & \textbf{7.7} & \textbf{5.6} & \textbf{8.2} & \textbf{203.3} & 28.8 & 19.8 & 60.0 & \textbf{6.8} & \textbf{8.4} & \textbf{5.6} & \textbf{10.0} & \textbf{46.2} & \textbf{47.0} \\
\cline{1-19} 
\midrule
\multirow{4}[1]{*}{\begin{sideways}\scriptsize{ CelebA-HQ}\end{sideways}} 
& \multicolumn{4}{l|}{DeepPrivacy} & 13.1 & 4.8 & 4.3 & 10.8 & 293.9 & 30.4 & 24.2 & 113.0 & 7.0 & 8.7 & 6.7 & 10.1 & 41.0 & 45.5 \\
& \multicolumn{4}{l|}{CIAGAN} & 14.9 & 10.3 & 4.6 & 8.6 & 365.6 & 79.2 & 35.6 & 91.5 & 8.7 & 13.7 & 7.5 & 13.4 & 38.0 & 44.4 \\
& \multicolumn{4}{l|}{RePaint} & 9.9 & \textbf{3.0} & 4.6 & 7.5 & 249.1 & \textbf{22.7} & 29.7 & 95.9 & 6.2 & 8.0 & 5.5 & 8.0 & 50.4 & 55.8 \\
& \multicolumn{4}{l|}{Ours} 
& \textbf{6.7} & 3.4 & \textbf{3.3} & \textbf{4.3} & \textbf{196.0} & 25.1 & \textbf{20.0} & \textbf{59.9} & \textbf{5.6} & \textbf{6.2} & \textbf{4.6} & \textbf{5.9} & \textbf{61.9} & \textbf{59.6} \\
\bottomrule
\end{tabular}
}
\vspace{-10pt}
\end{table*}

\begin{table}[t]
\centering
\small
\caption{Usability of de-identified datasets for the training of task-specific models (facial landmark detection on WFLW).}
\vspace{-6pt}
\label{tab:compare-utility-training}
\resizebox{\linewidth}{!}{
\begin{tabular}{l|ccccccc}
\toprule
\multicolumn{1}{c|}{\multirow{2}{*}{Methods}} & \multicolumn{7}{c}{Normalized Mean Error (NME) $\downarrow$} \\
\multicolumn{1}{c|}{} & all & pose & illu & occ & blur & mu & exp \\ \midrule
Original & .039 & .068 & .039 & .047 & .045 & .038 & .043\\ \hline
DeepPrivacy & .058 & .100 & .057 & .072 & .066 & .060 & .066\\
CIAGAN & .055 & .087 & .054 & .064 & .061 & .053 & .060\\
Ours & \textbf{.047} & \textbf{.079} & \textbf{.047} & \textbf{.056} & \textbf{.054} & \textbf{.046} & \textbf{.050}\\ \bottomrule
\end{tabular}
}\vspace{-4pt}
\end{table}

\noindent\textbf{Utility corruption in anonymized images.}
Our method demonstrates superior utility preservation compared to others across datasets (Table \ref{tab:compare-utility}). We highlight our approach's excellence through qualitative comparison (Figure \ref{fig:teaser} and \ref{fig:qualities}). DeepPrivacy lacks facial attribute preservation, exhibiting bias towards smiles and youth. CIAGAN bears heavy artifacts; Password yields blurry and easily re-identifiable outcomes. RePaint excels with in-distribution faces (RePaint is trained on CelebA-HQ thus has improved performance on the same dataset in Table \ref{tab:compare-utility}), but it fails elsewhere and doesn't retain original attributes. For challenging scenarios, like heavy occlusion (\eg, masks), CIAGAN and DeepPrivacy falter, unlike our effective face-swapping model.


\noindent\textbf{Usability of anonymized images as training data.
}
\label{sec:util_training}
We have demonstrated utility attribute non-corruption by comparing performance of pretrained task-specific models on obfuscated versus original data. Now, we advance toward the initial motivation of data anonymization for new solutions, evaluating how utility networks trained from scratch on anonymized data perform on real, unseen samples. Ideally, these privacy-preprocessed models should match performance of those trained on original, non-obfuscated data. Taking facial landmark detection as an example on the WFLW dataset \cite{wayne2018lab} (98 landmarks per image), we split data into training/testing sets (7,500/2,500) and generate obfuscated training data using mentioned methods (test data remains unaltered). We use an HRNetv2-W18 model \cite{wang2021hrnetv2} for the task, trained for 60 epochs with Adam optimizer \cite{kingma2014adam} ($\beta_1=0$, $\beta_2=0.999$), learning rate $10^{-4}$, and batch size 64. Table \ref{tab:compare-utility-training} shows models on obfuscated data perform worse (higher NME of facial landmarks) than the one on original data. Our anonymized data model demonstrates the smallest accuracy drop, confirming higher utility preservation for downstream tasks while maintaining privacy.

\subsection{Ablation Study}
Here we demonstrate the impact of different transformation models, identity and utility experts.

\noindent\textbf{Impact of transformation models on re-identifiability.}
As justified in Section~\ref{sec:method} and experimentally measured in Table~\ref{tab:compare-invertibility}, $\psi_{\text{opp}}$ would suffer high re-identification, \ie we can recover the original ID using the opposite of transformed ID.
MLP-based transformations outperforms opposite transformation but VED-based transformations yield the best results in terms of de-identification and non-invertibility, confirming the superiority of our proposed solution.

\begin{table}[t]
\centering
\small
\caption{Re-identifiability of our ID transformation methods.}
\vspace{-6pt}
\label{tab:compare-invertibility}

\resizebox{\linewidth}{!}{
\begin{tabular}{l@{\hspace{0.2em}}l@{\hspace{0.15em}}l@{\hspace{0.15em}}r|cccc}
\toprule
\multicolumn{4}{c|}{\multirow{3}{*}{Methods}} & \multicolumn{4}{c}{TPR (\%) @ FPR=1e-3 $\downarrow$ (LFW data)} \\
\multicolumn{4}{c|}{} & \multicolumn{2}{c|}{Swapped} & \multicolumn{2}{c}{Inverted} \\
\multicolumn{4}{c|}{} & \footnotesize{FaceNet} & \multicolumn{1}{c|}{\footnotesize{Sph.Face}} & \footnotesize{FaceNet} & \footnotesize{Sph.Face} \\ \midrule
Ours 
& {\scriptsize (arc,} 
& {\scriptsize opp,} 
& {\scriptsize \O)}
& 0.63 & \multicolumn{1}{c|}{0.03} & 67.03 & 53.07\\
Ours 
& {\scriptsize (arc,} 
& {\scriptsize ved,} 
& {\scriptsize emo)} 
& 0.23 & \multicolumn{1}{c|}{\textbf{0.00}} & \textbf{12.03} & 7.10 \\
Ours 
& {\scriptsize (arc,} 
& {\scriptsize ved,} 
& {\scriptsize eye)} 
& 0.03 & \multicolumn{1}{c|}{0.03} & 13.03 & \textbf{6.43}\\
\hdashline
Ours 
& {\scriptsize (arc,} 
& {\scriptsize mlp,} 
& {\scriptsize eye)}  
& \textbf{0.00} & \multicolumn{1}{c|}{\textbf{0.00}} & 52.90 & 45.97  \\
Ours 
& {\scriptsize (arc,} 
& {\scriptsize mlp,} 
& {\scriptsize emo)}  
& 0.03 & \multicolumn{1}{c|}{\textbf{0.00}} & 49.77 & 45.97 \\
\hdashline
Ours 
& {\scriptsize (arc,} 
& {\scriptsize mlp,} 
& {\scriptsize \underline{mix})} 
& \textbf{0.00} & \multicolumn{1}{c|}{\textbf{0.00}} & 50.23 & 44.67\\

Ours 
& {\scriptsize (\underline{mix},} 
& {\scriptsize  mlp,} 
& {\scriptsize eye)}  & 0.07 & \multicolumn{1}{c|}{\textbf{0.00}} & 36.70 & 34.70 \\ \bottomrule
\end{tabular}
}
\vspace{-10pt}
\end{table}


\begin{table}[t]
\centering
\small
\caption{Effect of $\psi^\epsilon_{\boldsymbol{\cdot}}$ noise \wrt (re-)identifiability.}
\vspace{-6pt}
\label{tab:both-noise}
\resizebox{\linewidth}{!}{
\begin{tabular}{ll|cc|cc}
\toprule
\multicolumn{2}{c|}{Methods} & \multicolumn{4}{c}{TPR (\%) @ FPR=1e-3 $\downarrow$  (LFW data)} \\
\multicolumn{1}{c}{\multirow{2}{*}{Network}} & \multicolumn{1}{c|}{\multirow{2}{*}{Noise}} & \multicolumn{2}{c|}{Swapped} & \multicolumn{2}{c}{Inverted} \\
& & FaceNet & \multicolumn{1}{c|}{Sph.Face} & FaceNet & Sph.Face \\ \midrule

\multicolumn{1}{c}{\multirow{3}{*}{MLP}} & $\beta=0.0$ & \textbf{0.00} & \textbf{0.00} & 52.90 & 45.97 \\
 & $\beta=0.5$ & 0.40 & \textbf{0.00} & 23.20 & 20.80 \\
 & $\beta=0.9$ & 5.90 & 2.50 & \textbf{3.07} & \textbf{1.30} \\ 
\hdashline
\multicolumn{1}{c}{\multirow{3}{*}{VED}} & $\alpha=1.0$ & \textbf{0.03} & \multicolumn{1}{c|}{0.03} & 13.03 & 6.43 \\
 & $\alpha=2.0$ & 0.37 & \multicolumn{1}{c|}{\textbf{0.00}} & 7.43 & 3.20 \\
 & $\alpha=3.0$ & 0.37 & \multicolumn{1}{c|}{0.10} & \textbf{5.37} & \textbf{2.23} \\ 
\bottomrule
\end{tabular}
}
\vspace{-10pt}
\end{table}

The introduction of stochastic operations in alignment with $\epsilon$-LDP further strengthen the solution. 
As shown in Table \ref{tab:both-noise}, the higher the amount of $\beta$ or $\alpha$ noise introduced (\ie, the lower $\epsilon$), the more robust to attacks the method becomes, but the lower the original de-identification rate (the noisier the data, the harder it is to synthesize an ID that maximizes obfuscation). This negative impact is however better mitigated by the proposed VED.
We provide further insights in supplementary material.


\noindent\textbf{Efeects of using multiple ID extractors.}
As shown in Table~\ref{tab:compare-invertibility}, MLP-based transformations relying on multiple identity extractors, \ie, ``Ours ($\text{mix}, \text{mlp}, \text{eye}$)", perform better than versions with only one ID expert. We attribute the increased robustness to the combined knowledge of the two algorithms which capture more varied ID-related features that are then obfuscated. 
\section{Conclusion and Discussion}

We introduced \textit{Disguise}, a privacy-enhancing face de-identification model that ensures both depicted people's privacy and image usability. Our experiments demonstrate its effectiveness in pre-processing sensitive data for inference or training. Rooted in privacy and mixture-of-experts theory, it outperforms prior methods in re-identification robustness and utility preservation.

\noindent\textbf{Limitations.}
Note that our model is tailored for face obfuscation and does not address other identity-revealing visual attributes (\eg, distinctive glasses, haircuts, backgrounds). Broader ID-extracting methods like $H_\mathcal{Z}$ \cite{bhanu2017deep} could potentially handle this. Additionally, \textit{Disguise} might benefit from multi-objective learning research \cite{desideri2012multiple,momma2022multi} to optimize cases where identity and utility features overlap.


\section{Acknowledgments}
Z. Cai and M. Asif were supported in part by AFOSR award FA9550-21-1-0330 and ONR award N00014-19-1-2264.

%
\bibliography{aaai24}

\begin{thebibliography}{77}
\providecommand{\natexlab}[1]{#1}

\bibitem[{HIP(2003)}]{HIPAA}
 2003.
\newblock Health Insurance Portability and Accountability Act.
\newblock U.S. Department of Health and Human Services.
\newblock 45 CFR Parts 160, 162, and 164.

\bibitem[{CCP(2018)}]{CCPA}
 2018.
\newblock California Consumer Privacy Act.
\newblock California Legislative Information.
\newblock Cal. Civ. Code \S1798.100 et seq.

\bibitem[{PIP(2021)}]{PIPL}
 2021.
\newblock Personal Information Protection Law.
\newblock National People's Congress of the People's Republic of China.

\bibitem[{noa(2022)}]{noauthor_insightface_2022}
 2022.
\newblock {InsightFace}: {2D} and {3D} {Face} {Analysis} {Project}.
\newblock \url{https://github.com/deepinsight/insightface}.

\bibitem[{Abadi et~al.(2016)Abadi, Chu, Goodfellow, McMahan, Mironov, Talwar, and Zhang}]{abadi2016deep}
Abadi, M.; Chu, A.; Goodfellow, I.; McMahan, H.~B.; Mironov, I.; Talwar, K.; and Zhang, L. 2016.
\newblock Deep learning with differential privacy.
\newblock In \emph{ACM CCS}.

\bibitem[{Abdelrahman et~al.(2022)Abdelrahman, Hempel, Khalifa, and Al-Hamadi}]{abdelrahman2022l2cs}
Abdelrahman, A.~A.; Hempel, T.; Khalifa, A.; and Al-Hamadi, A. 2022.
\newblock L2CS-Net: Fine-Grained Gaze Estimation in Unconstrained Environments.
\newblock \emph{arXiv:2203.03339}.

\bibitem[{Agarwal, Chattopadhyay, and Wang(2021)}]{agarwal2021privacy}
Agarwal, A.; Chattopadhyay, P.; and Wang, L. 2021.
\newblock Privacy preservation through facial de-identification with simultaneous emotion preservation.
\newblock \emph{SIVP}, 15(5).

\bibitem[{Barattin et~al.(2023)Barattin, Tzelepis, Patras, and Sebe}]{barattin2023attribute}
Barattin, S.; Tzelepis, C.; Patras, I.; and Sebe, N. 2023.
\newblock Attribute-preserving Face Dataset Anonymization via Latent Code Optimization.
\newblock In \emph{CVPR}.

\bibitem[{Bhanu, Kumar et~al.(2017)}]{bhanu2017deep}
Bhanu, B.; Kumar, A.; et~al. 2017.
\newblock \emph{Deep learning for biometrics}, volume~7.
\newblock Springer.

\bibitem[{Boyle, Edwards, and Greenberg(2000)}]{boyle2000effects}
Boyle, M.; Edwards, C.; and Greenberg, S. 2000.
\newblock The effects of filtered video on awareness and privacy.
\newblock In \emph{CSCW}.

\bibitem[{Cao et~al.(2021)Cao, Liu, Wen, Xie, and Song}]{cao2021personalized}
Cao, J.; Liu, B.; Wen, Y.; Xie, R.; and Song, L. 2021.
\newblock Personalized and Invertible Face De-identification by Disentangled Identity Information Manipulation.
\newblock In \emph{ICCV}.

\bibitem[{Cao et~al.(2018)Cao, Shen, Xie, Parkhi, and Zisserman}]{Cao18}
Cao, Q.; Shen, L.; Xie, W.; Parkhi, O.~M.; and Zisserman, A. 2018.
\newblock {VGGFace2}: A dataset for recognising faces across pose and age.
\newblock In \emph{International Conference on Automatic Face and Gesture Recognition}.

\bibitem[{Chen et~al.(2021)Chen, Chen, Yu, and Lu}]{chen2021perceptual}
Chen, J.-W.; Chen, L.-J.; Yu, C.-M.; and Lu, C.-S. 2021.
\newblock Perceptual Indistinguishability-Net (PI-Net): Facial image obfuscation with manipulable semantics.
\newblock In \emph{CVPR}.

\bibitem[{Chen et~al.(2020)Chen, Chen, Ni, and Ge}]{chen2020simswap}
Chen, R.; Chen, X.; Ni, B.; and Ge, Y. 2020.
\newblock Simswap: An efficient framework for high fidelity face swapping.
\newblock In \emph{ACM International Conference on Multimedia}.

\bibitem[{Croft, Sack, and Shi(2021)}]{croft2021obfuscation}
Croft, W.~L.; Sack, J.-R.; and Shi, W. 2021.
\newblock Obfuscation of images via differential privacy: From facial images to general images.
\newblock \emph{P2PNA}, 14(3).

\bibitem[{Dai et~al.(2021)Dai, Li, Liu, Tong, and Duan}]{dai2021generalizable}
Dai, Y.; Li, X.; Liu, J.; Tong, Z.; and Duan, L.-Y. 2021.
\newblock Generalizable person re-identification with relevance-aware mixture of experts.
\newblock In \emph{CVPR}.

\bibitem[{Deng et~al.(2020)Deng, Guo, Ververas, Kotsia, and Zafeiriou}]{deng2020retinaface}
Deng, J.; Guo, J.; Ververas, E.; Kotsia, I.; and Zafeiriou, S. 2020.
\newblock Retinaface: Single-shot multi-level face localisation in the wild.
\newblock In \emph{CVPR}.

\bibitem[{Deng et~al.(2019)Deng, Guo, Xue, and Zafeiriou}]{deng2019arcface}
Deng, J.; Guo, J.; Xue, N.; and Zafeiriou, S. 2019.
\newblock ArcFace: Additive Angular Margin Loss for Deep Face Recognition.
\newblock In \emph{CVPR}.

\bibitem[{D{\'e}sid{\'e}ri(2012)}]{desideri2012multiple}
D{\'e}sid{\'e}ri, J.-A. 2012.
\newblock Multiple-gradient descent algorithm (MGDA) for multiobjective optimization.
\newblock \emph{Comptes Rendus Mathematique}, 350(5-6).

\bibitem[{Duchi, Jordan, and Wainwright(2013)}]{duchi2013local}
Duchi, J.~C.; Jordan, M.~I.; and Wainwright, M.~J. 2013.
\newblock Local privacy and statistical minimax rates.
\newblock In \emph{FOCS}. IEEE.

\bibitem[{Dwork, Roth et~al.(2014)}]{dwork2014algorithmic}
Dwork, C.; Roth, A.; et~al. 2014.
\newblock The algorithmic foundations of differential privacy.
\newblock \emph{Foundations and Trends in Theoretical Computer Science}, 9(3--4).

\bibitem[{Fredrikson, Jha, and Ristenpart(2015)}]{fredrikson2015model}
Fredrikson, M.; Jha, S.; and Ristenpart, T. 2015.
\newblock Model inversion attacks that exploit confidence information and basic countermeasures.
\newblock In \emph{ACM CCS}.

\bibitem[{Frome et~al.(2009)Frome, Cheung, Abdulkader, Zennaro, Wu, Bissacco, Adam, Neven, and Vincent}]{frome2009large}
Frome, A.; Cheung, G.; Abdulkader, A.; Zennaro, M.; Wu, B.; Bissacco, A.; Adam, H.; Neven, H.; and Vincent, L. 2009.
\newblock Large-scale privacy protection in google street view.
\newblock In \emph{ICCV}.

\bibitem[{Gross et~al.(2005)Gross, Airoldi, Malin, and Sweeney}]{gross2005integrating}
Gross, R.; Airoldi, E.; Malin, B.; and Sweeney, L. 2005.
\newblock Integrating utility into face de-identification.
\newblock In \emph{International Workshop on Privacy Enhancing Technologies}. Springer.

\bibitem[{Gross et~al.(2009)Gross, Sweeney, Cohn, Torre, and Baker}]{gross2009face}
Gross, R.; Sweeney, L.; Cohn, J.; Torre, F. d.~l.; and Baker, S. 2009.
\newblock Face de-identification.
\newblock In \emph{Protecting privacy in video surveillance}. Springer.

\bibitem[{Gross et~al.(2006)Gross, Sweeney, De~la Torre, and Baker}]{gross2006model}
Gross, R.; Sweeney, L.; De~la Torre, F.; and Baker, S. 2006.
\newblock Model-based face de-identification.
\newblock In \emph{CVPR workshop}.

\bibitem[{Gu et~al.(2020)Gu, Luo, Ryoo, and Lee}]{gu2020password}
Gu, X.; Luo, W.; Ryoo, M.~S.; and Lee, Y.~J. 2020.
\newblock Password-conditioned anonymization and deanonymization with face identity transformers.
\newblock In \emph{ECCV}. Springer.

\bibitem[{Hempel, Abdelrahman, and Al-Hamadi(2022)}]{hempel20226d}
Hempel, T.; Abdelrahman, A.~A.; and Al-Hamadi, A. 2022.
\newblock 6D Rotation Representation For Unconstrained Head Pose Estimation.
\newblock \emph{arXiv:2202.12555}.

\bibitem[{Huang et~al.(2008)Huang, Mattar, Berg, and Learned-Miller}]{huang2008labeled}
Huang, G.~B.; Mattar, M.; Berg, T.; and Learned-Miller, E. 2008.
\newblock Labeled faces in the wild: A database for studying face recognition in unconstrained environments.
\newblock In \emph{Workshop on faces in 'Real-Life' Images}.

\bibitem[{Hukkel{\aa}s and Lindseth(2023)}]{hukkelaas2023deepprivacy2}
Hukkel{\aa}s, H.; and Lindseth, F. 2023.
\newblock Deepprivacy2: Towards realistic full-body anonymization.
\newblock In \emph{WACV}.

\bibitem[{Hukkel{\aa}s, Mester, and Lindseth(2019)}]{hukkelaas2019deepprivacy}
Hukkel{\aa}s, H.; Mester, R.; and Lindseth, F. 2019.
\newblock Deepprivacy: A generative adversarial network for face anonymization.
\newblock In \emph{ISVC}. Springer.

\bibitem[{Karras et~al.(2017)Karras, Aila, Laine, and Lehtinen}]{karras2017progressive}
Karras, T.; Aila, T.; Laine, S.; and Lehtinen, J. 2017.
\newblock Progressive growing of gans for improved quality, stability, and variation.
\newblock \emph{arXiv:1710.10196}.

\bibitem[{Kellnhofer et~al.(2019)Kellnhofer, Recasens, Stent, Matusik, and Torralba}]{kellnhofer2019gaze360}
Kellnhofer, P.; Recasens, A.; Stent, S.; Matusik, W.; and Torralba, A. 2019.
\newblock Gaze360: Physically unconstrained gaze estimation in the wild.
\newblock In \emph{ICCV}.

\bibitem[{Kim, Jain, and Liu(2022)}]{kim2022adaface}
Kim, M.; Jain, A.~K.; and Liu, X. 2022.
\newblock AdaFace: Quality Adaptive Margin for Face Recognition.
\newblock In \emph{CVPR}.

\bibitem[{King(2009)}]{dlib09}
King, D.~E. 2009.
\newblock Dlib-ml: A Machine Learning Toolkit.
\newblock \emph{Journal of Machine Learning Research}, 10.

\bibitem[{Kingma and Ba(2014)}]{kingma2014adam}
Kingma, D.~P.; and Ba, J. 2014.
\newblock Adam: A method for stochastic optimization.
\newblock \emph{arXiv:1412.6980}.

\bibitem[{Kingma, Salimans, and Welling(2015)}]{kingma2015variational}
Kingma, D.~P.; Salimans, T.; and Welling, M. 2015.
\newblock Variational dropout and the local reparameterization trick.
\newblock \emph{NeurIPS}, 28.

\bibitem[{Kingma and Welling(2013)}]{kingma2013auto}
Kingma, D.~P.; and Welling, M. 2013.
\newblock Auto-encoding variational bayes.
\newblock \emph{arXiv:1312.6114}.

\bibitem[{Korshunov and Ebrahimi(2013)}]{korshunov2013using}
Korshunov, P.; and Ebrahimi, T. 2013.
\newblock Using warping for privacy protection in video surveillance.
\newblock In \emph{DSP}, 1--6. IEEE.

\bibitem[{Li et~al.(2023)Li, Wang, Zhao, Dong, and Tan}]{li2023riddle}
Li, D.; Wang, W.; Zhao, K.; Dong, J.; and Tan, T. 2023.
\newblock RiDDLE: Reversible and Diversified De-identification with Latent Encryptor.
\newblock In \emph{CVPR}.

\bibitem[{Li et~al.(2020)Li, Bao, Yang, Chen, and Wen}]{li2019faceshifter}
Li, L.; Bao, J.; Yang, H.; Chen, D.; and Wen, F. 2020.
\newblock Faceshifter: Towards high fidelity and occlusion aware face swapping.
\newblock In \emph{CVPR}.

\bibitem[{Li and Clifton(2021)}]{li2021differentially}
Li, T.; and Clifton, C. 2021.
\newblock Differentially private imaging via latent space manipulation.
\newblock In \emph{SP}.

\bibitem[{Li and Lin(2019)}]{li2019anonymousnet}
Li, T.; and Lin, L. 2019.
\newblock Anonymousnet: Natural face de-identification with measurable privacy.
\newblock In \emph{CVPR workshop}.

\bibitem[{Liu et~al.(2021)Liu, Ding, Xue, Zhu, Ye, Song, and Zhou}]{liu2021dp}
Liu, B.; Ding, M.; Xue, H.; Zhu, T.; Ye, D.; Song, L.; and Zhou, W. 2021.
\newblock Dp-image: differential privacy for image data in feature space.
\newblock \emph{arXiv:2103.07073}.

\bibitem[{Liu et~al.(2017)Liu, Wen, Yu, Li, Raj, and Song}]{Liu_2017_CVPR}
Liu, W.; Wen, Y.; Yu, Z.; Li, M.; Raj, B.; and Song, L. 2017.
\newblock SphereFace: Deep Hypersphere Embedding for Face Recognition.
\newblock In \emph{CVPR}.

\bibitem[{Lugmayr et~al.(2022)Lugmayr, Danelljan, Romero, Yu, Timofte, and Van~Gool}]{lugmayr2022repaint}
Lugmayr, A.; Danelljan, M.; Romero, A.; Yu, F.; Timofte, R.; and Van~Gool, L. 2022.
\newblock Repaint: Inpainting using denoising diffusion probabilistic models.
\newblock In \emph{CVPR}.

\bibitem[{Masoudnia and Ebrahimpour(2014)}]{masoudnia2014mixture}
Masoudnia, S.; and Ebrahimpour, R. 2014.
\newblock Mixture of experts: a literature survey.
\newblock \emph{ARTR}, 42(2).

\bibitem[{Maximov, Elezi, and Leal-Taix{\'e}(2020)}]{maximov2020ciagan}
Maximov, M.; Elezi, I.; and Leal-Taix{\'e}, L. 2020.
\newblock Ciagan: Conditional identity anonymization generative adversarial networks.
\newblock In \emph{CVPR}.

\bibitem[{Miller and Uyar(1996)}]{miller1996mixture}
Miller, D.~J.; and Uyar, H. 1996.
\newblock A mixture of experts classifier with learning based on both labelled and unlabelled data.
\newblock \emph{NeurIPS}, 9.

\bibitem[{Momma, Dong, and Liu(2022)}]{momma2022multi}
Momma, M.; Dong, C.; and Liu, J. 2022.
\newblock A multi-objective/multi-task learning framework induced by Pareto stationarity.
\newblock In \emph{ICML}. PMLR.

\bibitem[{Neustaedter, Greenberg, and Boyle(2006)}]{neustaedter2006blur}
Neustaedter, C.; Greenberg, S.; and Boyle, M. 2006.
\newblock Blur filtration fails to preserve privacy for home-based video conferencing.
\newblock \emph{TOCHI}, 13(1).

\bibitem[{Newton, Sweeney, and Malin(2005)}]{newton2005preserving}
Newton, E.~M.; Sweeney, L.; and Malin, B. 2005.
\newblock Preserving privacy by de-identifying face images.
\newblock \emph{TKDE}, 17(2).

\bibitem[{Nirkin, Keller, and Hassner(2019)}]{nirkin2019fsgan}
Nirkin, Y.; Keller, Y.; and Hassner, T. 2019.
\newblock Fsgan: Subject agnostic face swapping and reenactment.
\newblock In \emph{ICCV}.

\bibitem[{Padilla-L{\'o}pez, Chaaraoui, and Fl{\'o}rez-Revuelta(2015)}]{padilla2015visual}
Padilla-L{\'o}pez, J.~R.; Chaaraoui, A.~A.; and Fl{\'o}rez-Revuelta, F. 2015.
\newblock Visual privacy protection methods: A survey.
\newblock \emph{Expert Systems with Applications}, 42(9).

\bibitem[{Perov et~al.(2020)Perov, Gao, Chervoniy, Liu, Marangonda, Um{\'e}, Dpfks, Facenheim, RP, Jiang et~al.}]{perov2020deepfacelab}
Perov, I.; Gao, D.; Chervoniy, N.; Liu, K.; Marangonda, S.; Um{\'e}, C.; Dpfks, M.; Facenheim, C.~S.; RP, L.; Jiang, J.; et~al. 2020.
\newblock DeepFaceLab: Integrated, flexible and extensible face-swapping framework.
\newblock \emph{arXiv:2005.05535}.

\bibitem[{Proen{\c{c}}a(2021)}]{proencca2021uu}
Proen{\c{c}}a, H. 2021.
\newblock The uu-net: Reversible face de-identification for visual surveillance video footage.
\newblock \emph{TCSVT}, 32(2).

\bibitem[{Qiu et~al.(2022)Qiu, Niu, Song, Ma, Al-Dhelaan, and Al-Dhelaan}]{qiu2022novel}
Qiu, Y.; Niu, Z.; Song, B.; Ma, T.; Al-Dhelaan, A.; and Al-Dhelaan, M. 2022.
\newblock A Novel Generative Model for Face Privacy Protection in Video Surveillance with Utility Maintenance.
\newblock \emph{Applied Sciences}, 12(14): 6962.

\bibitem[{Savchenko(2022)}]{savchenko2022video}
Savchenko, A.~V. 2022.
\newblock Video-Based Frame-Level Facial Analysis of Affective Behavior on Mobile Devices Using EfficientNets.
\newblock In \emph{CVPR}.

\bibitem[{Schroff, Kalenichenko, and Philbin(2015)}]{schroff2015facenet}
Schroff, F.; Kalenichenko, D.; and Philbin, J. 2015.
\newblock FaceNet: A unified embedding for face recognition and clustering.
\newblock In \emph{CVPR}.

\bibitem[{Sener and Koltun(2018)}]{sener2018multi}
Sener, O.; and Koltun, V. 2018.
\newblock Multi-task learning as multi-objective optimization.
\newblock \emph{NeurIPS}, 31.

\bibitem[{Serengil and Ozpinar(2021)}]{serengil2021lightface}
Serengil, S.~I.; and Ozpinar, A. 2021.
\newblock HyperExtended LightFace: A Facial Attribute Analysis Framework.
\newblock In \emph{ICEET}.

\bibitem[{Terhorst et~al.(2020)Terhorst, Kolf, Damer, Kirchbuchner, and Kuijper}]{terhorst2020ser}
Terhorst, P.; Kolf, J.~N.; Damer, N.; Kirchbuchner, F.; and Kuijper, A. 2020.
\newblock SER-FIQ: Unsupervised estimation of face image quality based on stochastic embedding robustness.
\newblock In \emph{CVPR}.

\bibitem[{T{\"o}lle et~al.(2022)T{\"o}lle, K{\"o}the, Andr{\'e}, Meder, and Engelhardt}]{tolle2022content}
T{\"o}lle, M.; K{\"o}the, U.; Andr{\'e}, F.; Meder, B.; and Engelhardt, S. 2022.
\newblock Content-Aware Differential Privacy with Conditional Invertible Neural Networks.
\newblock In \emph{DeCaF}. Springer.

\bibitem[{Voigt and Von~dem Bussche(2017)}]{voigt2017eu}
Voigt, P.; and Von~dem Bussche, A. 2017.
\newblock The eu general data protection regulation (gdpr).
\newblock \emph{A Practical Guide, 1st Ed.}

\bibitem[{Wang et~al.(2021)Wang, Sun, Cheng, Jiang, Deng, Zhao, Liu, Mu, Tan, Wang, Liu, and Xiao}]{wang2021hrnetv2}
Wang, J.; Sun, K.; Cheng, T.; Jiang, B.; Deng, C.; Zhao, Y.; Liu, D.; Mu, Y.; Tan, M.; Wang, X.; Liu, W.; and Xiao, B. 2021.
\newblock Deep High-Resolution Representation Learning for Visual Recognition.
\newblock \emph{TPAMI}, 43(10).

\bibitem[{Wang et~al.(2018)Wang, Liu, Zhu, Tao, Kautz, and Catanzaro}]{wang2018pix2pixHD}
Wang, T.-C.; Liu, M.-Y.; Zhu, J.-Y.; Tao, A.; Kautz, J.; and Catanzaro, B. 2018.
\newblock High-Resolution Image Synthesis and Semantic Manipulation with Conditional GANs.
\newblock In \emph{CVPR}.

\bibitem[{Wang, Si, and Wu(2015)}]{wang2015regression}
Wang, Y.; Si, C.; and Wu, X. 2015.
\newblock Regression model fitting under differential privacy and model inversion attack.
\newblock In \emph{IJCAI}.

\bibitem[{Wen et~al.(2022)Wen, Liu, Ding, Xie, and Song}]{wen2022identitydp}
Wen, Y.; Liu, B.; Ding, M.; Xie, R.; and Song, L. 2022.
\newblock Identitydp: Differential private identification protection for face images.
\newblock \emph{Neurocomputing}, 501.

\bibitem[{Wen et~al.(2021)Wen, Lin, Wang, and Xu}]{wen2021distract}
Wen, Z.; Lin, W.; Wang, T.; and Xu, G. 2021.
\newblock Distract your attention: multi-head cross attention network for facial expression recognition.
\newblock \emph{arXiv:2109.07270}.

\bibitem[{Westerlund(2019)}]{westerlund2019emergence}
Westerlund, M. 2019.
\newblock The emergence of deepfake technology: A review.
\newblock \emph{TIM Review}, 9(11).

\bibitem[{Wu et~al.(2018)Wu, Qian, Yang, Wang, Cai, and Zhou}]{wayne2018lab}
Wu, W.; Qian, C.; Yang, S.; Wang, Q.; Cai, Y.; and Zhou, Q. 2018.
\newblock Look at Boundary: A Boundary-Aware Face Alignment Algorithm.
\newblock In \emph{CVPR}.

\bibitem[{Xu et~al.(2022)Xu, Deng, Wang, Jing, Pan, and He}]{xu2022high}
Xu, Y.; Deng, B.; Wang, J.; Jing, Y.; Pan, J.; and He, S. 2022.
\newblock High-resolution face swapping via latent semantics disentanglement.
\newblock In \emph{CVPR}.

\bibitem[{Yu et~al.(2020)Yu, Xue, Liu, Wang, Zhu, and Ding}]{yu2020gan}
Yu, J.; Xue, H.; Liu, B.; Wang, Y.; Zhu, S.; and Ding, M. 2020.
\newblock Gan-based differential private image privacy protection framework for the internet of multimedia things.
\newblock \emph{Sensors}, 21(1).

\bibitem[{Zhang et~al.(2020)Zhang, Park, Beeler, Bradley, Tang, and Hilliges}]{Zhang2020ETHXGaze}
Zhang, X.; Park, S.; Beeler, T.; Bradley, D.; Tang, S.; and Hilliges, O. 2020.
\newblock ETH-XGaze: A Large Scale Dataset for Gaze Estimation under Extreme Head Pose and Gaze Variation.
\newblock In \emph{ECCV}.

\bibitem[{Zhou and Pun(2020)}]{zhou2020personal}
Zhou, J.; and Pun, C.-M. 2020.
\newblock Personal privacy protection via irrelevant faces tracking and pixelation in video live streaming.
\newblock \emph{TIFS}, 16.

\bibitem[{Zhou and Gregson(2020)}]{zhou2020whenet}
Zhou, Y.; and Gregson, J. 2020.
\newblock Whenet: Real-time fine-grained estimation for wide range head pose.
\newblock \emph{arXiv:2005.10353}.

\bibitem[{Zhu et~al.(2021)Zhu, Li, Wang, Xu, and Sun}]{zhu2021one}
Zhu, Y.; Li, Q.; Wang, J.; Xu, C.-Z.; and Sun, Z. 2021.
\newblock One shot face swapping on megapixels.
\newblock In \emph{CVPR}.

\end{thebibliography}
\newpage
\clearpage
\label{sec:appendix}
\begin{strip} 
    \begin{center} 
        {\LARGE  \textbf{Disguise without Disruption: Utility-Preserving Face De-Identification \\ (Supplementary Material)}}
    \end{center} 
\end{strip} 
\setcounter{section}{0}
\renewcommand\thesection{\Alph{section}}


\section*{Summary}
In this supplementary material, we augment the main paper in the following aspects: 
\begin{itemize}
    \item We provide comprehensive implementation details of our framework \textit{Disguise};
    \item We conduct additional ablation studies covering various variations of our methods;
    \item We include extra qualitative comparisons with other techniques using images in real-world scenarios, encompassing medical settings and images containing multiple faces.
\end{itemize}

\section{Further Implementation Details}
\label{sec:implementation}

\noindent\textbf{Architecture.} We build our face-swapping model $g$ based on the architecture and training framework proposed in SimSwap \cite{chen2020simswap}. The identity-merging module uses a $\MLP_{\theta_z}$ with two layers and feature sizes of [1024, 1024, 512]. The identity-transforming $\MLP_{\theta_z}$, on the other hand, is a 3-layer network with feature sizes [512, 2048, 1024, 512]. 
The VED encoder consists of two dense layers of sizes [512, 1024, 1024], followed by two parallel layers of size [512] to predict the mean and variance in latent space. The VED decoder has three dense layers of sizes [512, 1024, 1024, 512]. 
We use $\tanh$ as final activation throughout the model.

\begin{figure*}[t]
    \centering
    \includegraphics[width=0.9\linewidth]{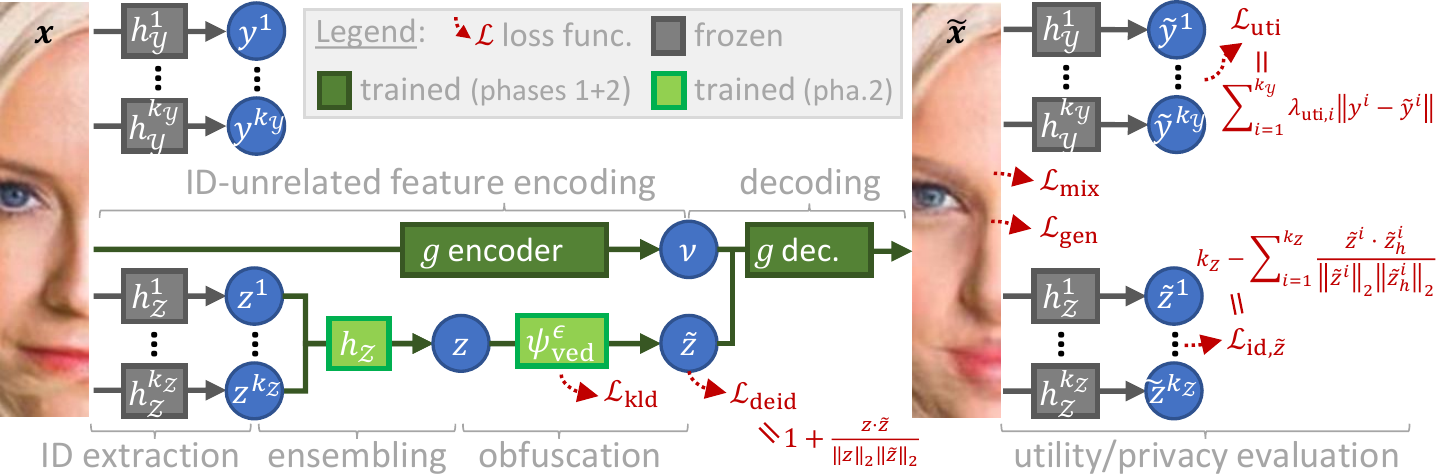}
    \caption{Detailed training pipeline of \textit{Disguise}, in supplement to Figure \ref{fig:architecure}. The proposed solution is end-to-end differentiable. However, in practice, to guide the optimization process, we train the network in two phases. Firstly, we train the face-swapping network (the branch marked in dark green); then in the second phase, we add the ID obfuscation branch (marked in light green) and the utility-guaranteeing module (the branch on top) to finetune the whole network.}
    \label{fig:training_phases}
\end{figure*}

\noindent\textbf{Training.} We apply the Adam optimizer \cite{kingma2014adam} with $\beta_1=0$ and $\beta_2=0.999$, learning rate $10^{-4}$, and a batch size of $4$. 
We train our pipeline in two stages: (1) we first pre-train the face-swapper $g$ according to \cite{chen2020simswap} for 1M iterations; (2) then we fine-tune it together with utility module and ID transformer for another 100k iterations. This is illustrated in Figure \ref{fig:training_phases}.

\noindent\textbf{Evaluation.} For the de-correlation evaluation presented in Figure \ref{fig:hist-4dist}(e), the MLP attacker networks consist of three layers of feature sizes [512, 2048, 1024, 512], tasked to reconstruct the original identity embedding from the obfuscated one extracted from the edited image. 
We train one attacker specific to each obfuscation method (CIAGAN \cite{maximov2020ciagan}, DeepPrivacy \cite{hukkelaas2019deepprivacy}, ours, \etc). 
We use Adam optimizer with a learning rate of $10^{-3}$ and a total epoch of 100 epochs. We trained the decoders on CelebA-HQ and evaluated them on LFW.



\section{More Ablation Studies and Analyses}
\label{sec:quantitative}

Complementing Ablation Study section in the main paper, we delve deeper into assessing how diverse ID extraction methods, ID transformation techniques, and utility experts can collectively influence the overall obfuscation pipeline.


\begin{figure}[!ht]
    \centering 
    \includegraphics[width=1\linewidth]{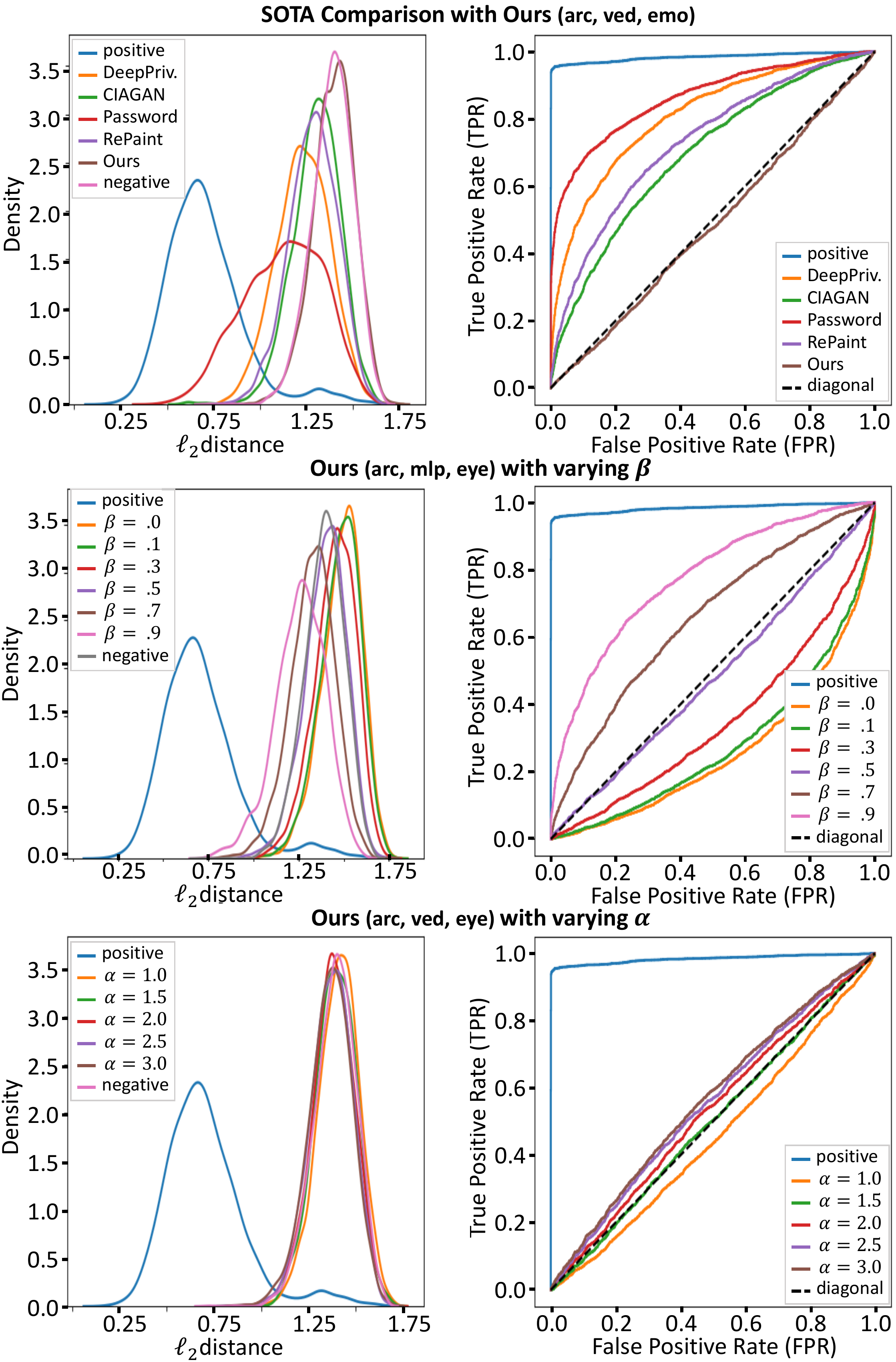}\\
    \caption{Left: Histogram of $\ell_2$ distances between positive, negative, and original-anonymized pairs from LFW set. Right: ROC curves of validation rate for images altered by various methods.
    }
    \label{fig:hist-all}
\end{figure}

\begin{table*}[t]
\centering
\small
\caption{Identification / validation rate ($\downarrow$, lower = better) and image quality evaluation ($\uparrow$, higher = better) over edited LFW data.}
\label{tab:compare-deid-ablation}

\begin{tabular}{l@{\hspace{0.2em}}l@{\hspace{0.15em}}l@{\hspace{0.15em}}r|cccc|c}
\toprule
\multicolumn{4}{c|}{\multirow{2}{*}{Methods}} & \multicolumn{4}{c|}{TPR (\%) @ FPR=$10^{-3}$ / Accuracy (\%) $\downarrow$} & FIQ $\uparrow$ \\
\multicolumn{4}{c|}{} & FaceNet & SphereFace & AdaFace & Average & SER \\ \midrule
\multicolumn{4}{l|}{Original} & 93.83 / 97.1 & 87.90 / 96.2 & 95.43 / 97.7 & 92.39 / 97.0 & 0.77 \\
\hline
Ours 
& {\scriptsize (arc,} 
& {\scriptsize opp,} 
& {\scriptsize \O)}  & 0.63 / 51.5 & 0.03 / 50.0 & 0.03 / 50.0 & 0.23 / 50.5 & 0.81 \\
Ours 
& {\scriptsize (arc,} 
& {\scriptsize ved,} 
& {\scriptsize emo)} 
& 0.23 / 50.1 & \textbf{0.00} / 50.0 & 0.07 / 50.0 & 0.10 / 50.0 & 0.90 \\
Ours 
& {\scriptsize (arc,} 
& {\scriptsize ved,} 
& {\scriptsize eye)} 
& 0.03 / 50.0 & 0.03 / 50.0 & \textbf{0.00} / 50.0 & 0.02 / 50.0 & 0.90 \\
\hdashline
Ours 
& {\scriptsize (arc,} 
& {\scriptsize mlp,} 
& {\scriptsize emo)} 
& 0.03 / 50.0 & \textbf{0.00} / 50.0 & 0.03 / 50.0 & 0.02 / 50.0 & 0.90 \\
Ours 
& {\scriptsize (arc,} 
& {\scriptsize mlp,} 
& {\scriptsize eye)} 
& \textbf{0.00} / 50.0 & \textbf{0.00} / 50.0 & 0.03 / 50.0 & \textbf{0.01 / 50.0} & 0.90 \\
Ours 
& {\scriptsize (arc,} 
& {\scriptsize mlp,} 
& {\scriptsize mix)} 
& \textbf{0.00} / 50.0 & \textbf{0.00} / 50.0 & 0.03 / 50.0 & \textbf{0.01 / 50.0} & 0.90 \\
\hdashline
Ours 
& {\scriptsize (ada,} 
& {\scriptsize mlp,} 
& {\scriptsize eye)} 
& 3.73 / 70.8 & 0.43 / 65.5 & \cellcolor{gray!33}  \color{white}NA \scriptsize{(\cf train/} & 2.08 / 68.2 & 0.87 \\
Ours 
& {\scriptsize (mix,} 
& {\scriptsize mlp,} 
& {\scriptsize eye)} 
& 0.07 / 50.0 & 0.00 / 50.0 & {\scriptsize \cellcolor{gray!33}  \color{white}eval overlap)} & 0.04 / 50.0 & \textbf{0.91} \\ \bottomrule
\end{tabular}
\end{table*}

\begin{table*}[t]
\centering
\small
\caption{Utility performance comparison of different versions of our methods over diverse downstream tasks on LFW dataset \cite{huang2008labeled}.}
\label{tab:compare-utility-ablation}
\resizebox{\linewidth}{!}{
\begin{tabular}{l@{\hspace{0.2em}}l@{\hspace{0.15em}}l@{\hspace{0.15em}}r|cccc|cccc|cccc|cc}
\toprule
\multicolumn{4}{c|}{\multirow{3}{*}{Methods}} & \multicolumn{8}{c|}{\begin{tabular}[c]{@{}c@{}}Facial landmarks (L2 pixel distance $\downarrow$)\end{tabular}} & \multicolumn{4}{c|}{\begin{tabular}[c]{@{}c@{}}Gaze estimation (MAE ° $\downarrow$)\end{tabular}} & \multicolumn{2}{c}{\begin{tabular}[c]{@{}c@{}}Emotion\end{tabular}} \\ \cline{5-18} 
\multicolumn{4}{c|}{} & \multicolumn{4}{c}{RetinaFace (5 points)} & \multicolumn{4}{c|}{Dlib (68 points)} & \multicolumn{2}{c}{L2CS Net} & \multicolumn{2}{c|}{ETH XGaze} & \multicolumn{2}{c}{(Accuracy \% $\uparrow$)} \\
\multicolumn{4}{c|}{} & All & Eyes & Nose & Mouth & All & Eyes & Nose & Mouth & Pitch & Yaw & Pitch & Yaw & DAN & DF \\ \midrule
Ours 
& {\scriptsize (arc,} 
& {\scriptsize opp,} 
& {\scriptsize \O)}
& \textbf{12.5} & \textbf{7.6} & \textbf{5.4} & \textbf{8.0} & \textbf{177.6} & \textbf{25.8} & \textbf{16.1} & \textbf{49.8} & 7.0 & 9.2 & 6.1 & 12.1 & 51.2 & 50.5 \\
Ours 
& {\scriptsize (arc,} 
& {\scriptsize ved,} 
& {\scriptsize emo)} 
& 12.7 & \textbf{7.6} & 5.7 & 8.1 & 187.6 & 26.5 & 18.8 & 55.6 & 7.4 & 10.8 & 6.1 & 13.1 & 58.8 & 49.8 \\
Ours 
& {\scriptsize (arc,} 
& {\scriptsize ved,} 
& {\scriptsize eye)} 
& 12.9 & 7.7 & 5.6 & 8.2 & 203.3 & 28.8 & 19.8 & 60.0 & 6.8 & 8.4 & 5.6 & 10.0 & 46.2 & 47.0 \\
\cdashline{1-18} 
Ours 
& {\scriptsize (arc,} 
& {\scriptsize mlp,} 
& {\scriptsize emo)} 
& 17.1 & 10.3 & 7.6 & 10.8 & 210.9 & 31.1 & 20.3 & 62.3 & 7.4 & 11.3 & 6.4 & 14.2 & \textbf{59.5} & \textbf{51.0} \\
Ours 
& {\scriptsize (arc,} 
& {\scriptsize mlp,} 
& {\scriptsize eye)} 
& 17.3 & 10.4 & 7.6 & 11.2 & 218.1 & 31.2 & 20.1 & 63.1 & 7.0 & 8.0 & 5.9 & 9.9 & 42.9 & 46.9 \\
Ours 
& {\scriptsize (arc,} 
& {\scriptsize mlp,} 
& {\scriptsize mix)} 
& 16.5 & 9.9 & 7.3 & 10.6 & 202.4 & 28.3 & 18.5 & 61.4 & 6.9 & 7.8 & 5.5 & 9.6 & \textbf{59.5} & 49.2 \\
\cdashline{1-18} 
Ours 
& {\scriptsize (arc,} 
& {\scriptsize mlp,} 
& {\scriptsize eye)}
& 17.3 & 10.4 & 7.6 & 11.2 & 218.1 & 31.2 & 20.1 & 63.1 & 7.0 & 8.0 & 5.9 & 9.9 & 42.9 & 46.9 \\
Ours 
& {\scriptsize (ada,} 
& {\scriptsize mlp,} 
& {\scriptsize eye)}
& 15.9 & 9.3 & 7.3 & 10.1 & 211.4 & 30.2 & 21.9 & 64.9 & 6.3 & \textbf{7.3} & \textbf{5.4} & 9.2 & 43.8 & 46.8 \\
Ours 
& {\scriptsize (sph,} 
& {\scriptsize mlp,} 
& {\scriptsize eye)}
& 14.9 & 8.9 & 6.5 & 9.5 & 205.2 & 28.7 & 21.2 & 60.5 & \textbf{6.1} & 7.5 & \textbf{5.4} & 10.3 & 47.4 & 49.4 \\
Ours 
& {\scriptsize (arc+ada,} 
& {\scriptsize mlp,} 
& {\scriptsize eye)}
& 15.7 & 9.3 & 6.9 & 10.3 & 202.8 & 27.0 & 18.7 & 62.1 & 7.2 & 7.8 & 6.2 & 9.7 & 42.8 & 46.9 \\
Ours 
& {\scriptsize (arc+sph,} 
& {\scriptsize mlp,} 
& {\scriptsize eye)}
& 19.7 & 11.7 & 8.4 & 12.8 & 230.1 & 34.0 & 22.7 & 66.0 & 6.8 & 7.4 & 5.8 & \textbf{9.0} & 39.1 & 45.9 \\ 
\bottomrule
\end{tabular}}
\end{table*}

\noindent\textbf{Impact of ID extraction models.}
Existing face swapping solutions \cite{chen2020simswap} also leverage out-of-the-box identification networks (\eg, ArcFace \cite{deng2019arcface} as the most common choice), but they do not provide any analysis on the possible bias that these pretrained methods may have and how such bias may impact the de-identification process, \eg, by improperly disentangling some facial features.

\begin{table}[t]
\centering
\small
\caption{Re-identification performance and invertibility of different proposed ID transformation methods. (mix1 indicates arc+ada, mix2 indicates arc+sph.)
}
\label{tab:supp-invertibility}
\resizebox{\linewidth}{!}{
\begin{tabular}{c|cccc}
\toprule
\multicolumn{1}{c|}{\multirow{3}{*}{Methods}} & \multicolumn{4}{c}{TPR (\%) @ FPR=1e-3 $\downarrow$} \\
\multicolumn{1}{c|}{} & \multicolumn{2}{c|}{Swapped} & \multicolumn{2}{c}{Inverted} \\
\multicolumn{1}{c|}{} & FaceNet & \multicolumn{1}{c|}{SphereFace} & FaceNet & SphereFace \\ \midrule
Ours \scriptsize(arc, mlp, eye) & \textbf{0.00} & \multicolumn{1}{c|}{\textbf{0.00}} & 52.90 & 45.97 \\
Ours \scriptsize(ada, mlp, eye) & 3.73 & \multicolumn{1}{c|}{0.43} & 49.77 & 42.63 \\
Ours \scriptsize(mix1, mlp, eye) & 0.07 & \multicolumn{1}{c|}{\textbf{0.00}} & 36.70 & \textbf{34.70} \\
Ours \scriptsize(sph, mlp, eye) & \textbf{0.00} &  \multicolumn{1}{c|}{\cellcolor{gray!33}}    & 68.70 & \cellcolor{gray!33} \\
Ours \scriptsize(mix2, mlp, eye) & \textbf{0.00} &  \multicolumn{1}{c|}{\multirow{-2}{*}{\cellcolor{gray!33}  \color{white}NA}}   & \textbf{31.93} & \multirow{-2}{*}{\cellcolor{gray!33}  \color{white}NA} \\

\bottomrule
\end{tabular}}
\end{table}

\begin{table}[t]
\centering
\small
\caption{Evaluation of ID transformation based on noise application only, in terms of de-identification and non-invertibility of the resulting images.}
\label{tab:supp_noise}
\resizebox{\linewidth}{!}{
\begin{tabular}{ll|cc|cc}
\toprule
\multicolumn{2}{c|}{Methods} & \multicolumn{4}{c}{TPR (\%) @ FPR=1e-3 $\downarrow$} \\
\multicolumn{1}{c}{\multirow{2}{*}{Net.}} & \multicolumn{1}{c|}{\multirow{2}{*}{Noise}} & \multicolumn{2}{c|}{Swapped} & \multicolumn{2}{c}{Inverted} \\
& & FaceNet & \multicolumn{1}{c|}{SphereFace} & FaceNet & SphereFace \\ \midrule

\multicolumn{1}{c}{\multirow{6}{*}{\O}} & $\beta = 0.25$ & 19.27 & \multicolumn{1}{c|}{12.87} & \textbf{1.27} & \textbf{0.30} \\
& $\beta = 0.5$ & 15.37 & \multicolumn{1}{c|}{9.90} & 1.37 & \textbf{0.30} \\
& $\beta = 1.0$ & 13.03 & \multicolumn{1}{c|}{7.73} & 1.47 & 0.43 \\
& $\beta = 2.0$ & 12.63 & \multicolumn{1}{c|}{6.23} & 1.93 & 0.57 \\
& $\beta = 4.0$ & \textbf{11.30} & \multicolumn{1}{c|}{6.50} & 1.47 & 0.47 \\
& $\beta = 8.0$ & \textbf{11.30} & \multicolumn{1}{c|}{\textbf{6.27}} & 2.23 & 0.50 \\ 
\bottomrule
\end{tabular}
}
\end{table}

To address this concern, we present our analyses in Tables \ref{tab:compare-deid-ablation}, \ref{tab:compare-utility-ablation}, and \ref{tab:supp-invertibility}. Our method can harness multiple ID extractors, thus we compare distinct versions of our solutions: employing ArcFace \cite{deng2019arcface}, AdaFace \cite{kim2022adaface}, SphereFace \cite{Liu_2017_CVPR}, or a fusion of these methods. Notably, we exclude the assessment on one ID extractor when it is utilized in the de-identification pipeline, \eg AdaFace in Table \ref{tab:compare-deid-ablation} last tow rows and SphereFace in Table \ref{tab:supp-invertibility} last tow rows, ensuring fairness.

Table \ref{tab:supp-invertibility} underscores that combining various ID extractors yields enhanced de-identification and non-invertibility. Particularly with AdaFace-based pipelines, this effect is evident. When solely used, AdaFace exhibits slight bias or performance limitations (vis-à-vis FaceNet \cite{schroff2015facenet} or SphereFace \cite{Liu_2017_CVPR} for re-identification), possibly due to missing biometric features, leading to higher re-identification rates post-obfuscation compared to other ID extractors. However, coupling AdaFace with an alternative ID method like ArcFace \cite{deng2019arcface} mitigates the re-identification rate effectively. Moreover, combining multiple identity extractors notably boosts resilience against inversion attacks (as evident in the last two columns of Table \ref{tab:supp-invertibility}), as anticipated from mixture-of-experts approaches.

Nonetheless, a trade-off between preserving privacy and utility remains observable. As Table \ref{tab:compare-utility-ablation} illustrates, solutions leveraging multiple ID extractors tend to exert a slightly greater impact on utility attributes, resulting in a minor accuracy dip for the designated downstream tasks. Navigating this trade-off and devising a solution that better disentangles identity and utility attributes—given their non-overlapping nature—remains an open challenge. Nevertheless, we believe that \textit{Disguise} represents a substantial stride forward in this regard (as evident from comparisons to state-of-the-art in both the main paper and this document).




\noindent\textbf{Impact of ID transformation models.}
Figure \ref{fig:hist-all} extends the analysis presented in Tables \ref{tab:compare-invertibility} and \ref{tab:both-noise}, highlighting the superiority of our VED-based obfuscation scheme compared to the other MLP-based proposed solution, as well as their superiority compared to prior art. 
The first row in Figure \ref{fig:hist-all} shows that compared to other methods, our VED-based model is further from the positive pairs both on the histogram and ROC curve, demonstrating the best de-identification ability. The second row shows that when we introduce more $\beta$ noise in our MLP model, both the histogram and ROC curve move closer towards the positive pairs. When $\beta=0.5$, our MLP model can de-identify facial images on which recognition model has performance close to random guess. For our VED model, when increasing the $\alpha$ noise, the histogram stays close to the negative pairs and the ROC curve stays close to the diagonal line, as shown in the third row. These results suggest that our VED model achieves the best de-identification while ensuring non re-identifiability. 

As a reminder, we define $\alpha$ and $\beta$ as inversely proportional to $\epsilon$, \cf $\alpha = \frac{\Delta\psi}{\epsilon}$ and $\beta = {\frac{\Delta\psi_{\text{mlp}}}{\epsilon}}$.
As a measure of privacy budget, the higher $\epsilon$ is fixed (\ie, the lower  $\alpha$ or $\beta$), the higher the privacy loss, \cf $\log{\mathrm{P}(\widetilde{z}|z)} - \log{\mathrm{P}(\widetilde{z}|z') \leq \epsilon}$ according to the formal definition in Problem Formulation subsection.
Local differential privacy (LDP) guarantees that an adversary observing $\widetilde{z}$ cannot determine with some degree of confidence if it comes from $z$ or $z'$. \Eg, $\epsilon = 0$ would mean zero confidence in linking a masked ID to a specific input one, as only noise would be transferred (\cf Laplacian noise with $\alpha = \frac{\Delta\psi}{\epsilon} = \infty$). 
To choose $\epsilon$ (and thus $\alpha$) adequately based on privacy budget, one should first estimate the sensitivity $\Delta$ of the processing function. Following standard practice \cite{liu2021dp,wen2022identitydp}, we measure the sensitivity of ours empirically: \eg, over LFW dataset, we obtain $\Delta\psi_{\text{ved}} = \sup_{z, z'}\left\|\psi(z) - \psi(z')\right\|_1 = 33.92$ (\eg, hence fixing $\alpha = 2$ means opting for a relative privacy budget equals to $\epsilon = 67.84$). 


We enhance the analysis presented in Table \ref{tab:both-noise} with additional insights from Table \ref{tab:supp_noise}. This new table illustrates the performance of the ID transformation scheme, which entails applying solely $\epsilon$-controlled Laplace noise to the features without employing additional neural networks for further vector obfuscation.
Comparatively, our proposed Multi-Layer Perceptron (MLP) and Variational Encoder-Decoder (VED) solutions distinctly elevate identity obfuscation beyond the effects of noise-only feature manipulation, as depicted in Table \ref{tab:supp_noise}. However, their continuous nature renders them more susceptible to re-identification risk, especially for similar privacy budgets. In cases of slight noise values, they could inadvertently map distinct inputs to a common fabricated identity. Despite this, we maintain the conviction that our VED-based approach strikes the most optimal balance between maximal de-identification and non-reidentifiability.

\noindent\textbf{Comparison with other noise-based ID tampering methods.}
Some other methods \cite{li2019anonymousnet,liu2021dp,li2021differentially,wen2022identitydp}, have been recently proposed to tackle de-identification of facial images by extracting identity features from the target data, altering it, and decoding it back into a similar but obfuscated image.
While we could not satisfyingly reproduce their results (no implementation has been released), we could approximate their solution using our own framework.
Indeed, most of these methods can be described as a subset of our modular solution, \ie, minus our main contributions. 
This is especially true for DP-Image \cite{liu2021dp} (not peer-reviewed yet) and IdentityDP \cite{wen2022identitydp} (published in August, the \nth{28}, 2022), which use an image encoder-decoder combined with an ID extractor \cite{deng2019arcface} and ID/image feature mixer, similar to ours. 
However, they do not provide our additional guarantees in terms of disentanglement of the facial attributes and preservation of the utility ones by using mixture-of-experts supervision. 
More importantly, they obfuscate the extracted ID vector (before injecting it with the residual image features and decoding it back into an image) only by adding Laplace noise to them. They do not leverage additional transformations in the ID latent space to ensure optimal de-identification, such as our MLP and VED neural functions.

\begin{figure*}[t!]
    \centering
    \includegraphics[width=1.0\textwidth]{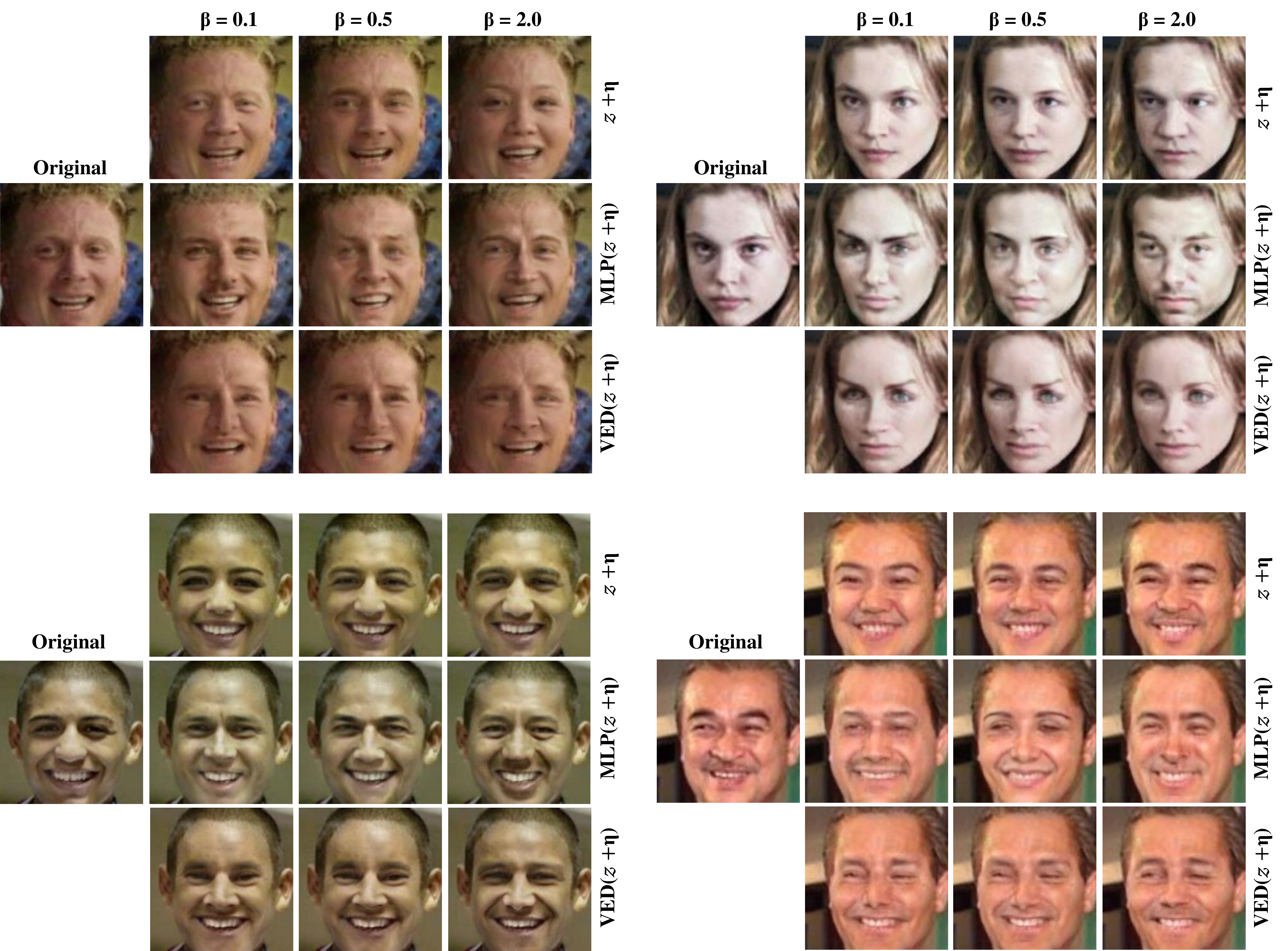}
    \caption{Comparison of different ID transformation functions in terms of their impact on the resulting obfuscated images. We compare (1) applying only Laplace noise to the extracted ID vectors (noted ``$z + \eta$" in the figure), (2) applying our proposed $\psi^\epsilon_{\text{mlp}}$, \ie, applying noise and our MLP (noted ``$\MLP(z + \eta)$"), or (3) applying our $\psi^\epsilon_{\text{ved}}$, \ie, applying noise and our VED (noted ``$\VED(z + \eta)$" here).}
    \label{fig:sup_qualinoise}
\end{figure*}

To highlight the impact of our proposed ID transformation functions and indirectly compare to these other solutions, we direct the readers to Figure \ref{fig:sup_qualinoise}.
For each original image, we display the results obtained by transforming the extracted ID features either after only adding Laplace-based noise to them (first row); after applying our proposed $\psi^\epsilon_{\text{mlp}}$, \ie, adding  Laplace noise and then passing the vector to our MLP optimized to ensure de-identification (second row); or after passing the vector to our VED $\psi^\epsilon_{\text{ved}}$, which also applies $\epsilon$-controlled noise to the data in its own latent space (third row).
For each solution, we provide several results with different privacy budgets ($\beta$ parameter, encompassing $\epsilon$).

We observe that applying only $\epsilon$-controlled noise to the ID vector results in images barely obfuscated (\eg, same nose/cheek/eyebrow shapes) compared to additionally using our proposed neural functions, for the same privacy budgets $\beta$. Furthermore, our VED-based solution provides better continuity in the obfuscated results \wrt $\beta$ compared to the other two variants. Such continuity makes choosing an adequate privacy budget much more intuitive and straightforward for users.


\noindent\textbf{Impact of utility experts.}
Our observations indicate that engaging in fine-tuning alongside utility experts yields notable enhancements in preserving performance for downstream tasks, as evidenced by the findings in Tables \ref{tab:compare-deid-ablation} and \ref{tab:compare-utility-ablation}. To delve into specifics, we note that in the gaze estimation task, the model fine-tuned with the inclusion of eye-related utility experts showcases the most minimal offset, denoting superior alignment. Similarly, the same pattern emerges in the context of emotion recognition, where the model fine-tuned with emotion-centric utility experts achieves optimal results. On the other hand, concerning facial landmarks, intriguing dynamics come to light. The model deprived of fine-tuning with emotion or gaze experts demonstrates the highest performance in this regard. This contrast highlights the existence of a trade-off phenomenon, indicating that distinct utility experts exert varying degrees of influence, necessitating a balanced consideration.

\section{More Qualitative Evaluation and Comparisons}
\label{sec:qualitative}

In this section, we provide additional qualitative results highlighting the quality of privacy and utility preservation provided by the proposed method, compared to the most popular face anonymization methods, such as blurring \cite{frome2009large}, pixelation \cite{zhou2020personal}, Password \cite{gu2020password}, CIAGAN \cite{maximov2020ciagan}, DeepPrivacy \cite{hukkelaas2019deepprivacy}, and Repaint \cite{lugmayr2022repaint}. We also exhibit the ability to directly run on images in the wild without the need of complex post-precessing, in contrast to \cite{li2021differentially,barattin2023attribute}. 

\begin{figure*}[t!]
    \centering
    \includegraphics[width=1.0\textwidth]{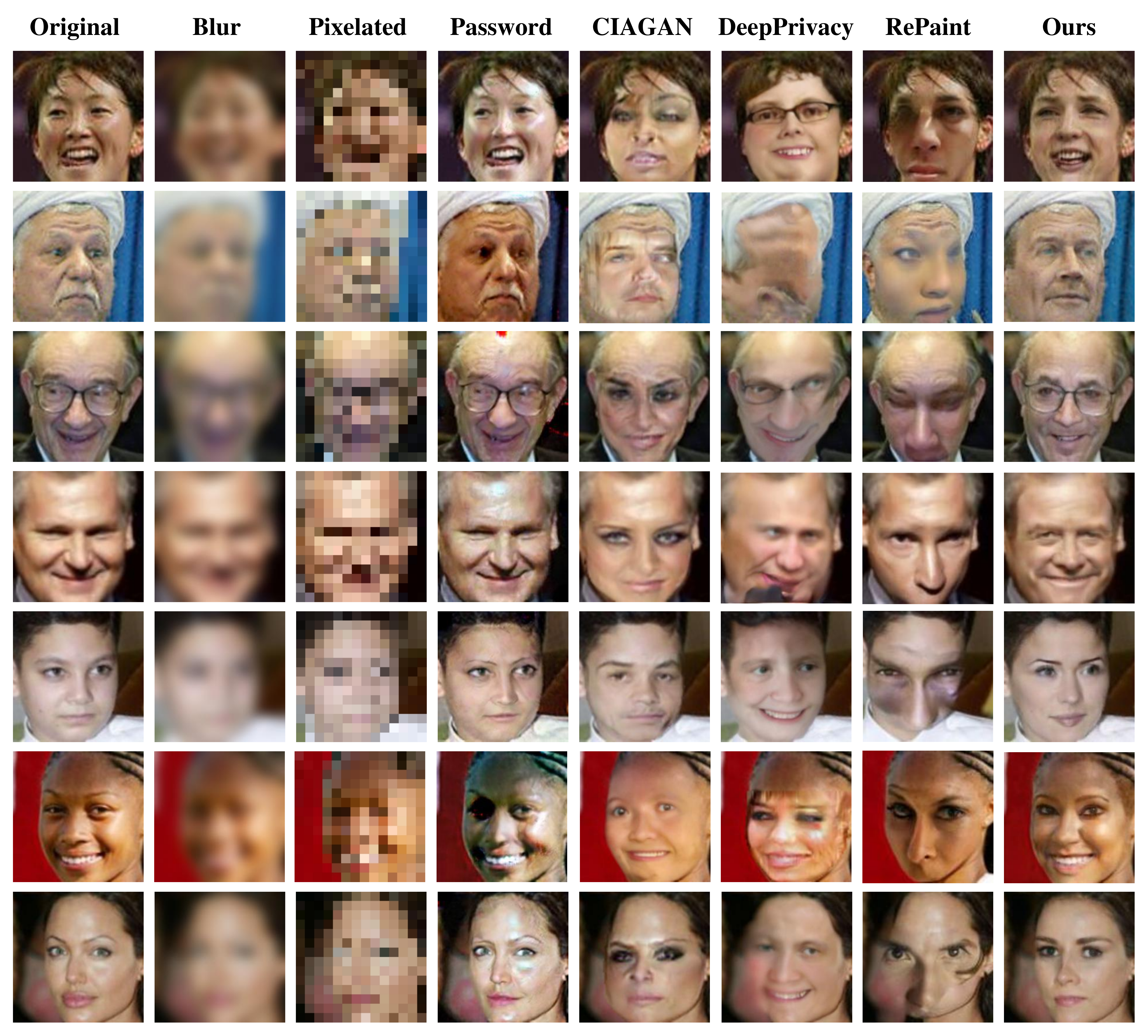}
    \caption{Qualitative results of face anonymization methods. Our method anonymizes images while maintaining utility and image quality.}
    \label{fig:sup_qualities}
\end{figure*}

\noindent\textbf{More qualitative comparisons on LFW.} Figure \ref{fig:sup_qualities} complements Figure \ref{fig:qualities} with more comparisons. Simpler methods, such as blurring and pixelation, provide effective anonymization, but the resulting facial images cannot be leveraged for downstream tasks. 
For the other deep-learning-based solutions, the observations are similar to those made \wrt Figure \ref{fig:qualities}. 
Password \cite{gu2020password} fails at removing identifying features; whereas RePaint \cite{lugmayr2022repaint} has difficulties hallucinating entire new faces, resulting in images too distorted to be useful. CIAGAN \cite{maximov2020ciagan} and DeepPrivacy \cite{hukkelaas2019deepprivacy} provide strong anonymization and pseudo-realistic results, but they still suffer from artifacts that can also impair usability (in terms of saliency and utility attributes).
Our method sometimes struggles with out-of-distribution images (\eg, facial images with lighting conditions hiding key features) causing some utility loss, but it overall yields realistic images sharing utility attributes with the original ones while successfully altering identifying traits (nose width, thickness of the eyebrows, shape of cheeks, \etc).



\begin{figure*}[t]
    \centering
    \includegraphics[width=.8\textwidth]{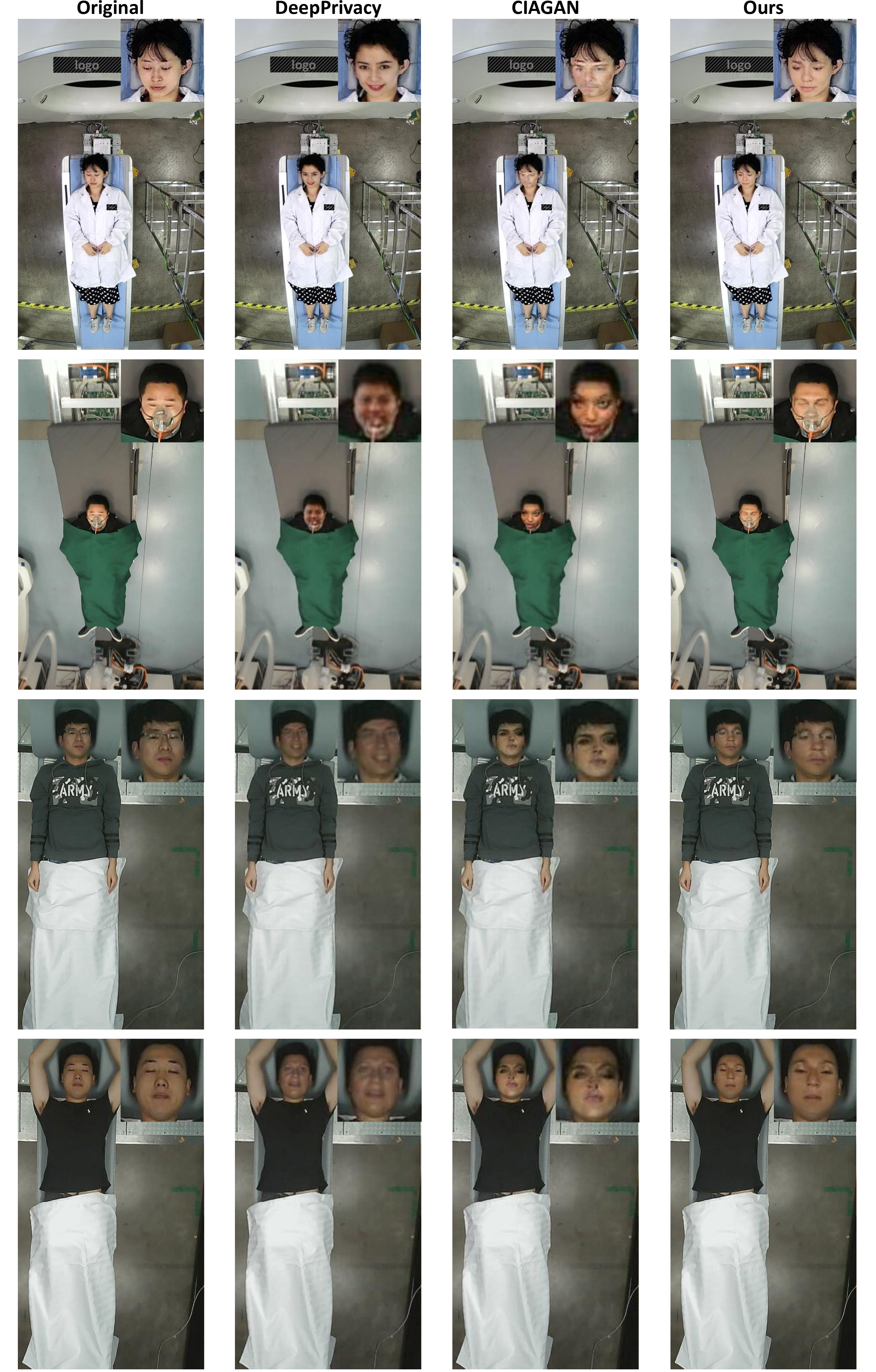}
    \caption{Qualitative results of face anonymization methods on images collected in medical settings.}
    \label{fig:sup_qualities_medical}
\end{figure*}

\noindent\textbf{Sensitive images taken in medical settings.} Closer to the target use-cases discussed in the Introduction, we also share additional results on sensitive images taken in medical settings (the images were taken, edited, and shared with the consent of the depicted volunteers), \cf Figure \ref{fig:sup_qualities_medical}. Once again, we note the superiority of the proposed method in terms of image quality and usability. For example, the gaze and facial expression are better preserved, and so are elements occluding the faces (oxygen mask, glasses). If the obfuscated data were to be used for training algorithms on face-focused tasks for medical environments, preserving such challenging non-facial features would be important to ensure the robustness of these methods after deployment. 

\begin{figure*}[t]
    \centering
    \includegraphics[width=1.0\textwidth]{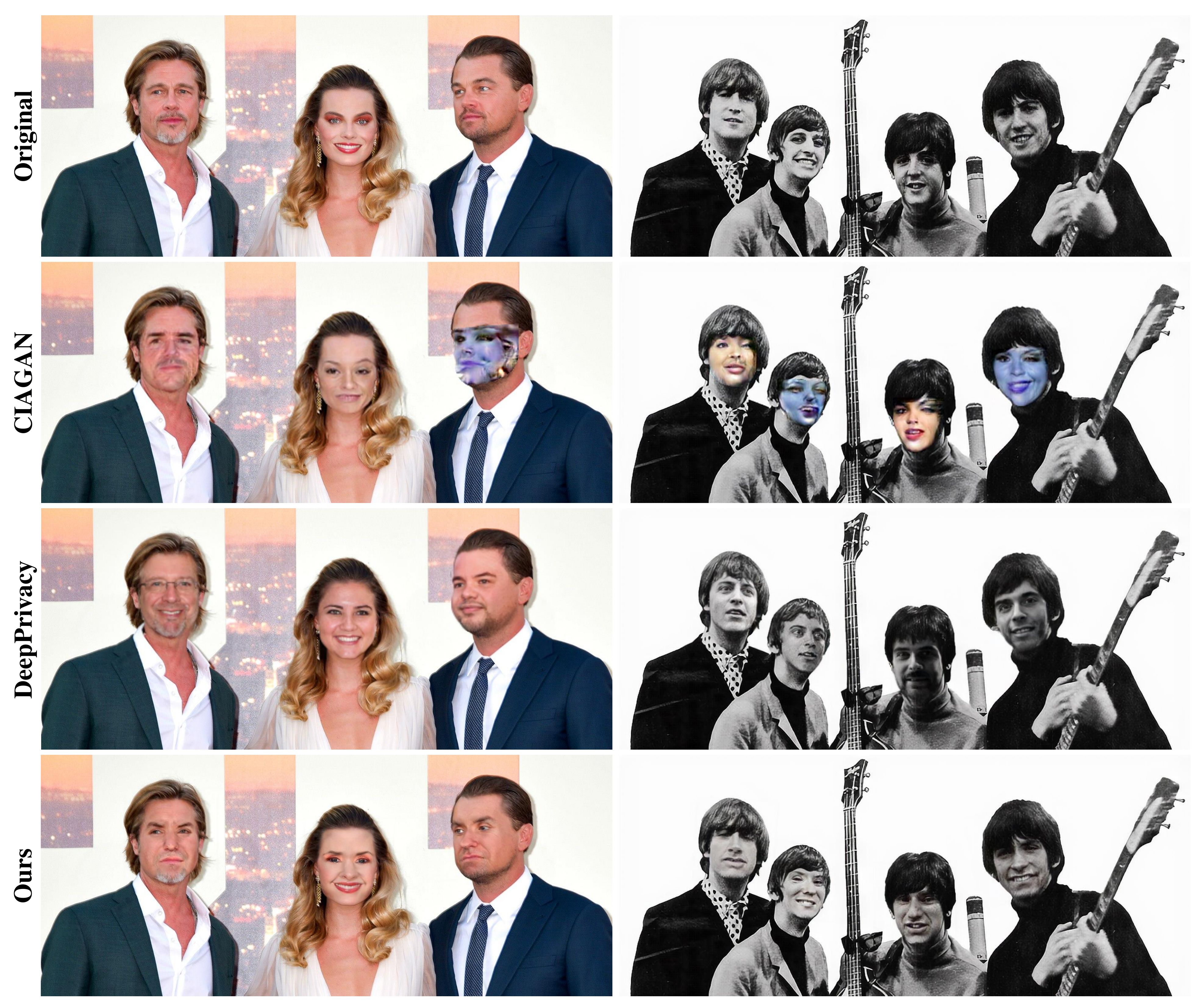}
    \caption{Qualitative results of face anonymization methods on images depicting multiple persons.}
    \label{fig:sup_qualities_group}
\end{figure*}

\noindent\textbf{Group photos with multiple faces.} Figure \ref{fig:sup_qualities_group} shows some results when applying the evaluated methods to group images, again highlighting the performance of our method compared to the state-of-the-art. \Eg, while DeepPrivacy is able to generate high-quality \textit{fake} faces, it does not preserve key utility attributes as well as our method (\eg, changing gaze directions or facial expressions, adding glasses, \etc). 

Note that for such group images or images showing more than just a face (\cf Figure \ref{fig:sup_qualities_group}), we first apply InsightFace \cite{noauthor_insightface_2022,deng2019arcface}, a face detection model, to obtain the region for each face present in the image; then we apply the de-identification methods to each corresponding crop separately; before merging everything back into the obfuscated image.

\end{document}